\pdfoutput=1

\documentclass[11pt]{article}

\usepackage{acl}
\usepackage{times}
\usepackage{latexsym}
\usepackage[T1]{fontenc}
\usepackage[utf8]{inputenc}
\usepackage{microtype}
\usepackage{inconsolata}

\usepackage[utf8]{inputenc} 
\usepackage[T1]{fontenc}    
\usepackage{hyperref}       
\usepackage{url}            
\usepackage{booktabs}       
\usepackage{amsfonts}       
\usepackage{nicefrac}       
\usepackage{microtype}      
\usepackage{xcolor}   
\usepackage{threeparttable}
\usepackage{listings} 
\usepackage{colortbl}
\usepackage{multicol}
\usepackage{multirow}
\usepackage{wrapfig}
\usepackage{graphicx}
\usepackage{tcolorbox}
\usepackage{float} 
\usepackage{amsmath}
\usepackage{enumitem}
\definecolor{darkpurple}{RGB}{160, 0, 160}
\definecolor{darkred}{RGB}{160, 0, 0}
\definecolor{darkblue}{RGB}{0, 0, 160}
\usepackage{algorithm}
\usepackage{algorithmic}
\usepackage{amsfonts}
\usepackage{caption}
\definecolor{softblue}{RGB}{100, 149, 237}
\usepackage{hyperref}  
\hypersetup{
    colorlinks,
    linkcolor=darkpurple,
    anchorcolor=blue,
    citecolor=softblue
}
\usepackage{graphicx}

\title{Disentangling Memory and Reasoning Ability in Large Language Models}

\author{
Mingyu Jin$^1$\quad
Weidi Luo$^2$\quad 
Sitao Cheng$^3$\quad 
Xinyi Wang$^3$\quad \\
\textbf{Wenyue Hua}$^3$\quad
\textbf{Ruixiang Tang}$^1$\quad
\textbf{William Wang}$^3$\quad 
\textbf{Yongfeng Zhang}$^{1}$\thanks{Corresponding Email: mingyu.jin@rutgers.edu yongfeng.zhang@rutgers.edu. }\\
  $^1$Rutgers University\;\;\;  $^2$The Ohio State University\;\;\;  $^3$University of California, Santa Barbara\;\;\; 
}

\begin{document}
\maketitle
\begin{abstract}

Large Language Models (LLMs) have demonstrated strong performance in handling complex tasks that require both extensive knowledge and reasoning abilities. However, the existing LLM inference pipeline operates as an opaque process without explicit separation between knowledge retrieval and reasoning steps, making the model’s decision-making process unclear and disorganized. Recent research has shown that this ambiguity will lead to issues such as knowledge forgetting, which significantly impact the reliability of LLMs. In this paper, we propose a novel language model inference paradigm that decomposes the complex inference process into two distinct and clear actions: \textbf{(1) memory recall}: which retrieves relevant knowledge in LLM, and \textbf{(2) reasoning}: which performs reasoning steps based on the recalled knowledge. To facilitate this decomposition, we introduce two special tokens \textbf{$\langle \text{memory} \rangle$} and \textbf{$\langle \text{reason} \rangle$}, guiding the model to distinguish between steps that require knowledge retrieval and those that involve reasoning.  Our experiment results show that this decomposition not only improves LLMs' performance among utility benchmarks but also enhances interpretability during the inference process, enabling users to identify sources of error and refine model responses effectively. The code is available at: https://github.com/MingyuJ666/Disentangling-Memory-and-Reasoning.
\end{abstract}

\section{Introduction}
 Recent advancements in Large Language Models (LLMs) have showcased their impressive inference capabilities in handling complex natural language tasks that require both extensive knowledge and sophisticated reasoning abilities \cite{openai2023gpt4,touvron2023LLaMA,wei2022emergent}. LLMs have demonstrated the ability to memorize vast amounts of knowledge, and techniques like \textit{Chain-of-Thought (CoT)}~\cite{wei2022chain}, \textit{Tree of thoughts (ToT)}~\cite{yao2024tree} have been developed to further enhance their inference abilities by decomposing complex problems into several simpler, single-step processes. These methods enable LLMs to tackle multi-step inference tasks more effectively by organizing the thought process into discrete, focused actions~\cite{feng2024towards,jin2024impact,wei2022chain,sun2025visual}.

Despite these advancements, existing inference frameworks often operate as an opaque process without explicitly separating knowledge retrieval and reasoning steps. This makes it unclear what specific knowledge the model utilizes and how it performs reasoning, leaving the decision-making process ambiguous. For complex, knowledge-intensive tasks, LLMs often struggle to effectively leverage their memory for inference~\cite{yang2023chatgpt, jin2024impact,cheng2024understandinginterplayparametriccontextual, liu2024mindstepbystep}. Such tasks typically require the ability to recall relevant knowledge for each reasoning step and then perform inference over that recalled memory~\cite{wangunderstanding}. The lack of structure in the output and the inefficient memory utilization can result in issues such as knowledge forgetting, where relevant information is lost across reasoning steps~\cite{chen2023llm}, which disrupts the logical flow, as well as hallucinations, where LLMs generate plausible yet incorrect information~\cite{xu2024tow,li2024deceptive}. These issues compromise the LLM's accuracy and reliability, posing serious risks in high-stakes applications like healthcare and finance~\cite{pham2024towards}.

Existing efforts to enhance inference in LLMs and address their challenges can be broadly classified into two main approaches: Memory-based Approaches: These methods focus on improving the recall and utilization of world knowledge that may not be stored in the model, such as leveraging Retrieval-Augmented Generation (RAG)~\cite{cai-etal-2019-skeleton, chen2024spiral}. The emphasis is on enabling models to access and use their outside knowledge more effectively. Reasoning-based Approaches: These techniques aim to improve the reasoning capabilities of models by Chain-of-Thought (CoT) reasoning~\cite{yang2023chatgpt, gao2024two, yu2024distilling} or introducing structured guidance in training such as planning tokens~\cite{wang2024guidinglanguagemodelreasoning, wang2024m2pt} to organize reasoning into discrete, interpretable steps. These methods enhance the ability of LLMs to handle complex reasoning tasks by embedding structural reasoning mechanisms into their parameters. Despite advancements in both categories, LLMs still struggle with tasks that require an intricate interplay of memory recall and logical reasoning~\cite{wangunderstanding}.

In this work, we propose a novel LLM inference paradigm that divides the complex inference process into two distinct components: memory and reasoning. Specifically, we generate itemized action responses for various question-answering datasets, categorizing each action as either memory or reasoning. Each action is then preceded by a special token, either $\langle \text{memory} \rangle$ or $\langle \text{reason} \rangle$, which acts as a control signal during training. The second step involves training an LLM using these modified outputs. By incorporating these learnable control tokens, the model is explicitly guided to distinguish between recalling relevant knowledge and performing reasoning steps. This structured guidance encourages the model to first use memory to retrieve the relevant information and then apply reasoning based on that memory to solve the task. Our approach not only introduces a new form of structured response generation but also establishes a novel framework for guiding LLMs to "think" systematically. This structured decomposition improves both the model's performance and the interpretability of its inference process.

Our experimental results demonstrate that the proposed decomposition improves performance and enhances the interpretability of the model’s inference process. Specifically, our method achieves accuracy of \textbf{78.6\%} and \textbf{78.0\%} on the StrategyQA dataset~\cite{geva2021strategyqa} using Qwen2.5-7B~\cite{qwen2} and LLaMA-3.1-8B~\cite{touvron2023LLaMA}, respectively. These results represent improvements of 1.2\% and 1.3\% over the planning-token fine-tuned baseline while remaining only 2.2\% below GPT-4o's performance. Remarkably, on the TruthfulQA dataset~\cite{lin2022truthfulqa}, LLaMA-3.1-8B enhanced by our algorithm outperforms GPT-4o with Chain of Thought prompting (85.4\%), achieving \textbf{86.6\%} accuracy. On average across three benchmark datasets, our method narrows the performance gap with the top-performing closed-source model, GPT-4o (using CoT prompting), to just 1.9\%. Furthermore, by analyzing the errors made by LLaMA-3.1-8B, we reveal that most issues stem from reasoning rather than deficiencies in the knowledge itself. This distinction sheds light on the primary sources of errors in the model’s outputs and enables targeted improvements. 


Our main contributions are as follows:

\begin{itemize}[leftmargin=*, itemsep=0pt] \item \textbf{New Inference Paradigm for LLMs:} We introduce a framework that decomposes inference in LLMs into \textbf{memory} and \textbf{reason} steps, guiding the model to separate knowledge retrieval from logical reasoning, thus enhancing performance and interpretability.
\item \textbf{Advancing Benchmark Performance:} Our model achieves competitive results, surpassing GPT-4o on TruthfulQA and closely matching GPT4-o on StrategyQA and CommonsenseQA, demonstrating the benefits of our approach.
\item \textbf{Empowering Transparency and Control:} Our framework enables transparent reasoning with labeled steps for memory and reasoning, allowing precise error analysis and model refinement.
\end{itemize}

\begin{figure*}[!th]
    \centering
    \includegraphics[width=1\linewidth]{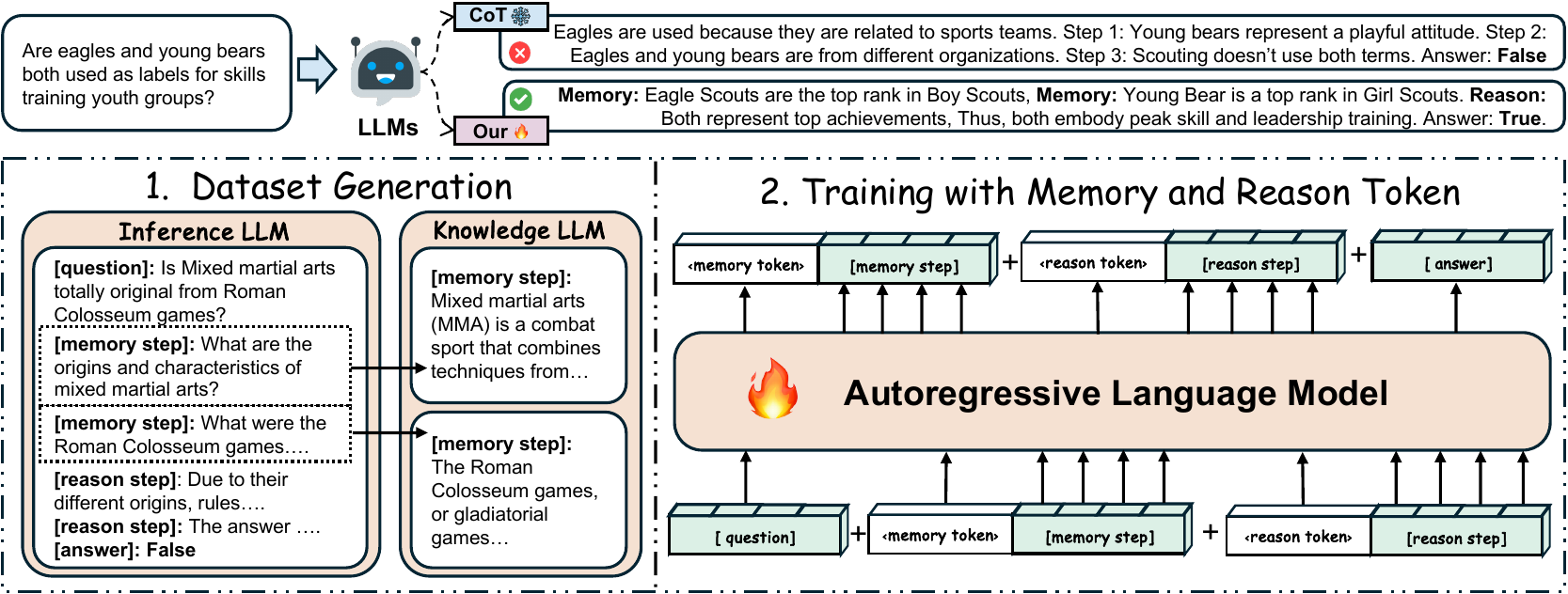}
    \caption{\textbf{Workflow. }We employ an LLM-based framework for data generation by two LLMs: an inference LLM that generates reasoning and memory steps, and a knowledge LLM that supplies the factual knowledge required for those memory steps. The generated data is annotated with two distinct special tokens: {$\langle \text{memory} \rangle$} and {$\langle \text{reason} \rangle$}, which are used for training the autoregressive language model alongside the question and answer.}
    \vspace{-0.8em}
    \label{fig:workflow}
\end{figure*}
\section{Method}
The workflow of our method can be divided into two stages: Data generation by decoupling memory and reasoning steps and training LLM with memory and reasoning tokens on generated data.

\subsection{ Data Generation with Decoupled Memory and Reasoning}
We introduce an LLM-based framework for response generation to generate memory (knowledge in LLM) and reasoning steps, consisting of an inference LLM and a knowledge LLM, as illustrated in Figure \ref{fig:workflow}. First, we use an inference LLM to generate Chain of Thought (CoT)~\citep{wei2022chain} inference steps, prompting it to mark steps that require factual knowledge as {$\langle \text{memory} \rangle$} and those requiring reasoning as {$\langle \text{reason} \rangle$}. To improve the quality of the memory steps, we further instruct the inference LLM to rephrase knowledge marked as {$\langle \text{memory} \rangle$} into questions, emphasizing its factual nature. For the example question in Figure~\ref{fig:workflow}, The inference LLM first retrieves relevant knowledge {$\langle \text{memory} \rangle$} about MMA and Roman Colosseum games, analyzes their relationship {$\langle \text{reason} \rangle$} and synthesizes this information to form a coherent judgment. By methodically aligning each step with its purpose, the LLM ensures that the conclusion—MMA is not "totally original" from the Colosseum games—reflects well-supported reasoning.  Next, a knowledge LLM answers the questions about factual knowledge generated by inference LLM, such as \textit{What are the
origins and characteristics of mixed martial arts?} and \textit{What were the
Roman Colosseum games?}. The answers to these questions are then substituted into the CoT inference steps.  This approach effectively decouples reasoning from knowledge, ensuring accuracy while maintaining high data quality. It enables the fine-tuning of LLMs by disentangling knowledge and reasoning during inference. We leverage this LLM-based framework between memory and reasoning steps to generate interpretable data, which can be used for the training stage.

\subsection{LLM Training with Memory and Reasoning Tokens}

At this stage, we train an LLM by incorporating intervened reasoning and memory processes as Figure \ref{fig:workflow}, guided by two special tokens: $\langle \text{reason} \rangle$, which represents reasoning with knowledge, and $\langle \text{memory} \rangle$, which signifies retrieved factual knowledge. These special tokens are designed to prompt the model to activate the necessary knowledge for reasoning, strengthening its inference capabilities and ultimately enhancing both interpretability and performance in complex inference tasks.
During training stage, each training instance $\mathcal{T}$ comprises the following components: (1) the question tokens $\mathcal{Q} = \{q_1, q_2, ..., q_{n_Q}\}$ where $n_Q$ is the question token length, (2) the step-by-step thinking process consists of intertwined memory and reasoning components, denoted as $M$ and $R$, where each $M$ is initiated by a special token $\langle \text{memory} \rangle$ followed by a sequence of tokens $K$ that represent retrieved factual knowledge: $\{\text{$\langle \text{memory} \rangle$}, k_1, k_2, ..., k_{n_K}\}$, and $R$ is initiated by a special token $\langle \text{Reason} \rangle$ followed by a sequence of tokens $S$ that represent the reasoning process: $\{\text{$\langle \text{reason} \rangle$}, s_1, s_2, ..., s_{n_S}\}$, and (3) the target answer generated after the completion of the memory retrieval and reasoning processes. The model is trained in a standard autoregressive manner using LoRA fine-tuning and the {$\langle \text{reason} \rangle$} and {$\langle \text{memory} \rangle$} are trainable out of vocabulary tokens. 

By structuring the input in this paradigm, the model learns to process and distinguish between retrieved knowledge and the reasoning steps required to generate the final answer. The inclusion of the $\langle \text{memory} \rangle$ and $\langle \text{reason} \rangle$ tokens facilitates the disentanglement of memory retrieval and reasoning processes, thereby enhancing the model's ability to produce coherent and accurate responses.

\section{Experiment}

\subsection{Experiment Setup}
\label{Setting Up}
\paragraph{Models.} In our experiments, we use  LLaMA-2-7B-chat-hf~\cite{touvron2023LLaMA}, LLaMA-3.1-8B-Instruct~\cite{dubey2024LLaMA}, and Qwen2.5-7B-Instruct~\cite{qwen2} as backbone models for training and test. GPT-4o serves as the inference and knowledge LLM to generate training data, while GPT-4o-mini is employed as the evaluator.

\paragraph{Datasets.} 
Our experiments are carried out on three data sets: StrategyQA~\cite{geva2021strategyqa} is a question-answer benchmark of 2,780 examples. Each example includes questions, supporting evidence, and answers. CommonsenseQA~\cite{talmor-etal-2019-commonsenseqa} contains 12,102 questions, each of which requires common sense knowledge to select the correct answer from four distractors. TruthfulQA~\cite{lin2022truthfulqa} evaluates the truthfulness of the responses to the language model, with 817 questions in 38 categories. We used the \textit{mc1\_targets} subset, which consists of single-choice questions with 4-5 answer choices. To prepare this dataset for training and testing, we labeled the answer options as A-E and shuffled the labels to avoid shortcuts in training. For StrategyQA and CommonsenseQA, we used their predefined training and testing set splits. For TruthfulQA, we split the data into an 8:2 ratio for training and testing.

\paragraph{Baselines.} In our experiments, we adopt zero shot (just input the question) and CoT prompting as our vanilla baseline for inference. For the fine-tuned baseline, we choose LoRA fine-tuning~\cite{hu2021loralowrankadaptationlarge} and Planning Tokens (LoRA+Prompt Tuning)~\cite{wang2024guidinglanguagemodelreasoning} to train and test on these three datasets to facilitate a comparative evaluation with our approach. In training, we use int8\_training to save GPU memory and accelerate.

\paragraph{Evaluation Metric.} We use accuracy (acc) to measure the model's performance on all datasets.

\subsection{Main Results}
Five main methods are being compared: Zero-shot, CoT (Chain-of-Thought), LoRA, Planning-token, and Ours (mentioned in Section \ref{Setting Up}).  The results in StrategyQA and CommonsenseQA benchmarks indicate that our algorithm consistently achieves higher scores across both benchmarks compared to other approaches, particularly in fine-tuned models. For instance, in StrategyQA, Our method enhanced LLaMA-3.1-8B achieved a score of 78.0\%, outperforming CoT at 69.4\% and Planning-token at 76.7\%. Similarly, in CommonsenseQA, Our method enhanced LLaMA-3.1-8B scores 82.3\%, compared to CoT’s 70.6\% and Planning-token’s 76.9\%, suggesting the effectiveness of our algorithm in improving LLMs' performance.

For the TruthfulQA dataset, we achieved a significant breakthrough; the LLaMA-3.1-8B enhanced by our algorithm (86.6\%) even outperforms GPT-4o in both zero-shot (84.8\%) and CoT settings (85.4\%), which is remarkable. GPT-4 sometimes gets misled by these options in this dataset, but our model effectively handles these challenges. Our model first considers relevant knowledge and then uses it in reasoning, which proves highly effective on this dataset(as the appendix~\ref{appendix:case_study}, we include an analysis of both correct and incorrect examples). However, Qwen2.5-7B performed poorly on this dataset, achieving only 81.0\% in our algorithm, likely due to instruction tuning in Qwen2.5, resulting in average performance and unstable training. However, adding a CoT can decrease performance for some models in some datasets, which is also a phenomenon reported by \cite{sprague2024cot}.

\begin{table*}[ht]
\centering
\small 
\caption{Main Comparative Experiment Results.}
\label{tab:main_result}
\begin{tabular}{p{1.8cm}p{2.6cm}cccc} 
\toprule
\rowcolor[RGB]{234, 238, 234} \textbf{Methods} & \textbf{Models} &  \textbf{StrategyQA} & \textbf{CommonSenseQA} & \textbf{TruthfulQA} & \textbf{Average} \\
\midrule
\multicolumn{6}{c}{\textit{Vanilla}}   \\ \midrule

&  LLaMA-2-7B~ & 0.607 & 0.523 & 0.262 & 0.464 \\
\multirow{0}{*}{Zero-shot} &  LLaMA-2-13B~ & 0.613 & 0.530 & 0.378 & 0.507 \\
& LLaMA-3.1-8B~ & 0.659 & 0.635 & 0.616 & 0.637 \\
& LLaMA 3.1-70B~ & 0.796 & 0.765 & 0.793 & 0.785 \\
& Qwen 2.5-7B~ & 0.640 & 0.789 & 0.726 & 0.718 \\
& GPT-4o~ & 0.699 & 0.834 & 0.848 & 0.794 \\ \midrule

&  LLaMA-2-7B~ & - & - & - & - \\
\multirow{0}{*}{CoT} &  LLaMA-2-13B~ & 0.560 & 0.482 & 0.390 & 0.477 \\
& LLaMA-3.1-8B~ & 0.694 & 0.706 & 0.506 & 0.635 \\
& LLaMA 3.1-70B~ & 0.822 & 0.815 & 0.762 & 0.800 \\
& Qwen 2.5-7B~ & 0.696 & 0.784 & 0.567 & 0.682 \\
& GPT-4o~ & \textbf{0.808} & \textbf{0.865} & \underline{0.854} & 0.842 \\ \midrule

\multicolumn{6}{c}{\textit{Fine-tuned}}   \\\midrule

&  LLaMA-2-7B~ & 0.612 & 0.641 & 0.767 & 0.673 \\
&  LLaMA-2-13B~ & 0.696 & - & - & 0.696 \\
\multirow{-2}{*}{LoRA} & LLaMA-3.1-8B~ & 0.701 & 0.754 & 0.798 & 0.737 \\
& Qwen 2.5-7B~ & 0.691 & 0.775 & 0.725 & 0.730 \\
 \midrule
&  LLaMA-2-7B~ & 0.635 & 0.654 & 0.770 & 0.686 \\
&  LLaMA-2-13B~ & 0.715 & - & - & 0.715 \\
 \multirow{-2}{*}{Planning-token} & LLaMA-3.1-8B~ & 0.767 & 0.769 & 0.825 & 0.787 \\
& Qwen 2.5-7B~ & 0.774 & 0.801 & 0.762 & 0.779 \\ \midrule
&  LLaMA-2-7B~ & 0.706 & 0.711 & 0.786 & 0.734 \\
&  LLaMA-2-13B~ & 0.739 & - & - & 0.739 \\
 \multirow{-2}{*}{Ours} & LLaMA-3.1-8B~ & 0.780 & 0.823 & \textbf{0.866} & 0.823 \\
& Qwen 2.5-7B~ & \underline{0.786} & \underline{0.832} & 0.812 & 0.810 \\
\bottomrule
\end{tabular}
\end{table*}

\begin{table*}
    \centering
    \caption{Ablation study with LLaMA-3.1-8B and LLaMA-2-7B on three benchmarks.}
    \label{tab:ablation}
    \resizebox{1.6\columnwidth}{!}{
    \begin{tabular}{lcccc}
    \toprule
   \rowcolor[RGB]{234, 238, 234}  & \textbf{StrategyQA}  & \textbf{CommonsenseQA}  & \textbf{TruthfulQA} & \textbf{Average} \\
    \midrule
     LLaMA-2-7B         && \\ 
    ~~~~w Memory and Reason token & 0.706 & 0.711 & 0.786 & 0.734 \\
    ~~~~w Random token  & 0.644 & 0.651 & 0.708 & 0.668 \\
    LLaMA-3.1-8B         && \\ 
    ~~~~w Memory and Reason token & 0.780 & 0.823 & 0.866 & 0.823 \\
    ~~~~w Random token  & 0.759 & 0.795 & 0.840 & 0.798 \\
    \bottomrule
    \end{tabular}
    }
\end{table*}
\vspace{-0.5em}

\subsection{Ablation Study}
In the ablation study, we comprehensively investigate the effects of the impact of special token~\ref{section:ablation_study:impact_special_token} and the number of special tokens~\ref{section:ablation_study:impact_count}.
\subsubsection{Impact of Memory and Reason Tokens}
\label{section:ablation_study:impact_special_token}
The Ablation Experiment presents an ablation study comparing the performance of two versions of the LLaMA model (LLaMA-2-7B and LLaMA-3.1-8B) across three benchmarks: StrategyQA, CommonsenseQA, and TruthfulQA. The study examines the impact of using specific tokens ("Memory and Reason") vs.Random tokens on the model's performance. During training, we shuffled the allocation of {$\langle \text{reason} \rangle$} and {$\langle \text{memory} \rangle$} tokens and then observed the effects on training and testing performance. As expected, the overall performance declined (shown in Table~\ref{tab:ablation}), but the decline rate varied (from 2.1\% to 6.6\%), showing our approach's superiority in disentangling reason and memory.

\subsubsection{Impact of Special Token Count}
\label{section:ablation_study:impact_count}
In our training setup, we have two token types: {$\langle \text{reason} \rangle$} and {$\langle \text{memory} \rangle$}, and we will include a parameter representing the number of special tokens preceding each sentence. For example, A sentence might include three {$\langle \text{reason} \rangle$} tokens or four, like the question in Appendix~\ref{app:training_details:example}. Our experiments indicate that model performance reaches a higher point with around four to six special tokens (as Table~\ref{num1} and~\ref{num2}). This is likely because more tokens may lead to better performance for the LLM~\cite{levy-etal-2024-task}, as proved by previous research. We selected two LLMs to illustrate their performance (ACC) across different numbers of special tokens. 

Another important issue is knowledge distillation. We must ensure that the model's improvement is not due to knowledge distillation from the GPT-4 framework. Using the same inference steps, we compared the results of standard training with 0 reason and memory tokens and found that adding these tokens significantly improves performance. This indirectly confirms that the model's enhancement comes from algorithmic improvements rather than knowledge distillation.

\begin{table*}[t]
\setlength{\tabcolsep}{2pt}
\centering
\small
\begin{minipage}{0.44\textwidth}
    \centering
    \caption{Model Performance (ACC) by Number of Special Tokens in CommonsenseQA}
    \begin{tabular}{lcccc}
    \toprule
    \multicolumn{1}{c}{\textbf{Model Name}} & \multicolumn{4}{c}{\textbf{Number of Tokens}} \\ 
    \cmidrule(lr){2-5}
    & \textbf{0} & \textbf{2} & \textbf{4} & \textbf{6} \\
    \midrule
     LLaMA-2-7B & 0.682 & 0.704 & 0.710 & 0.711 \\
    LLaMA-3.1-8B & 0.783 & 0.816 & 0.823 & 0.820 \\
    Qwen2.5-7B & 0.799 & 0.813 & 0.832 & 0.813 \\
    \bottomrule
    \end{tabular}
    \label{num1}
\end{minipage}%
\hspace{50pt}
\begin{minipage}{0.44\textwidth}
    \centering
    \caption{Model Performance (ACC) by Number of Special Tokens in TruthfulQA}
    \begin{tabular}{lcccc}
    \toprule
    \multicolumn{1}{c}{\textbf{Model Name}} & \multicolumn{4}{c}{\textbf{Number of Tokens}} \\ 
    \cmidrule(lr){2-5}
    & \textbf{0} & \textbf{2} & \textbf{4} & \textbf{6} \\
    \midrule
     LLaMA-2-7B & 0.701 & 0.762 & 0.768 & 0.786 \\
    LLaMA-3.1-8B & 0.826 & 0.865 & 0.866 & 0.859 \\
    Qwen2.5-7B & 0.807 & 0.756 & 0.812 & 0.799 \\
    \bottomrule
    \end{tabular}
    \label{num2}
\end{minipage}%
\end{table*}

\subsection{Further Analysis}
In this section, we aim to analyze the decoupling effect~\ref{case_study:sample_analysis}, attention analysis~\ref{case_study:attention_analysis} of our method and error analysis of our method for ~\ref{case_study:error_analysis}.
\subsubsection{Decoupling Analysis}
\label{case_study:sample_analysis}
To validate the decoupling effect on memory and reasoning, we configure GPT-4o-mini as an evaluator (details in Appendix~\ref{appendix:case_study:Sample Analysis}), assessing whether steps labeled as "memory" entail factual knowledge and those labeled as "reasoning" represent reasoning processes on our three benchmarks. Then We use a structured, directive one-shot Chain of Thought (CoT) prompting method to prompt LLaMA-3.1-8B as the baseline that can also disentangle memory and reason step. This prompt setup is displayed in Appendix~\ref{appendix:case_study:Sample Analysis} in Figure~\ref{fig:one_shot_cot}.

\begin{figure*}[!h]
    \centering
    \begin{minipage}[t]{0.31\textwidth}
        \centering
        \includegraphics[width=\textwidth]{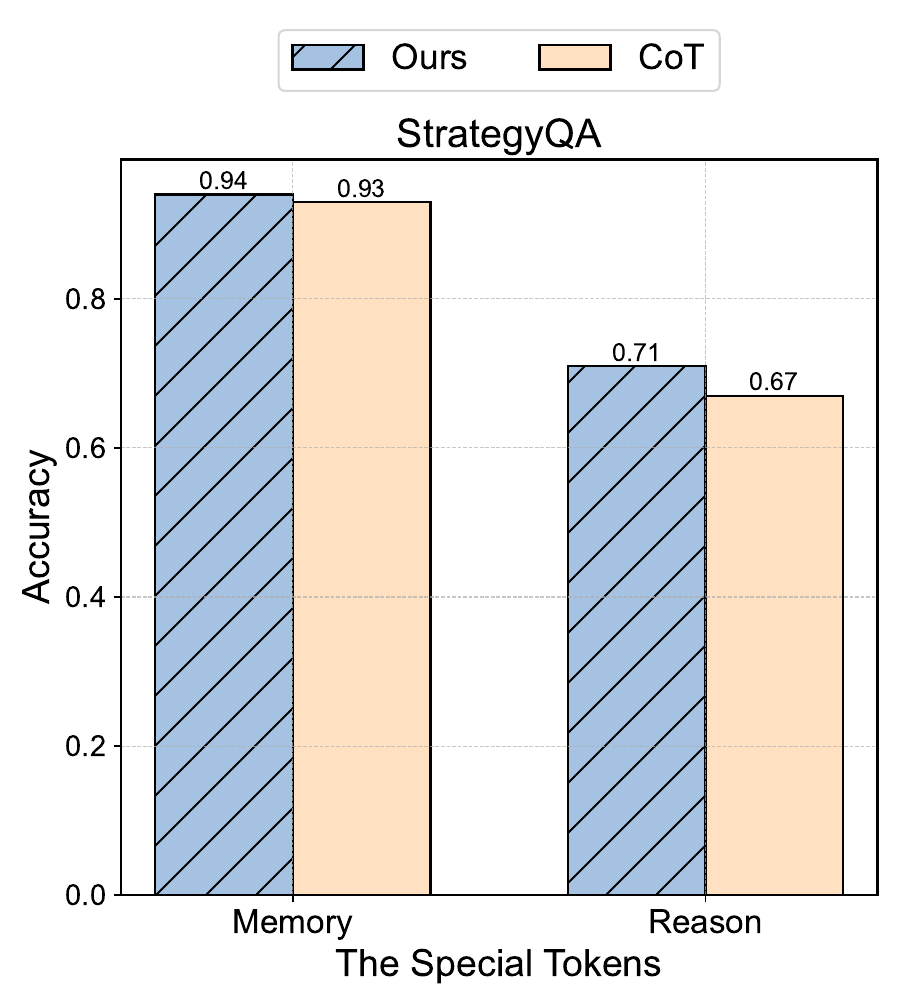}
    \end{minipage}
    \hfill
    \begin{minipage}[t]{0.31\textwidth}
        \centering
        \includegraphics[width=\textwidth]{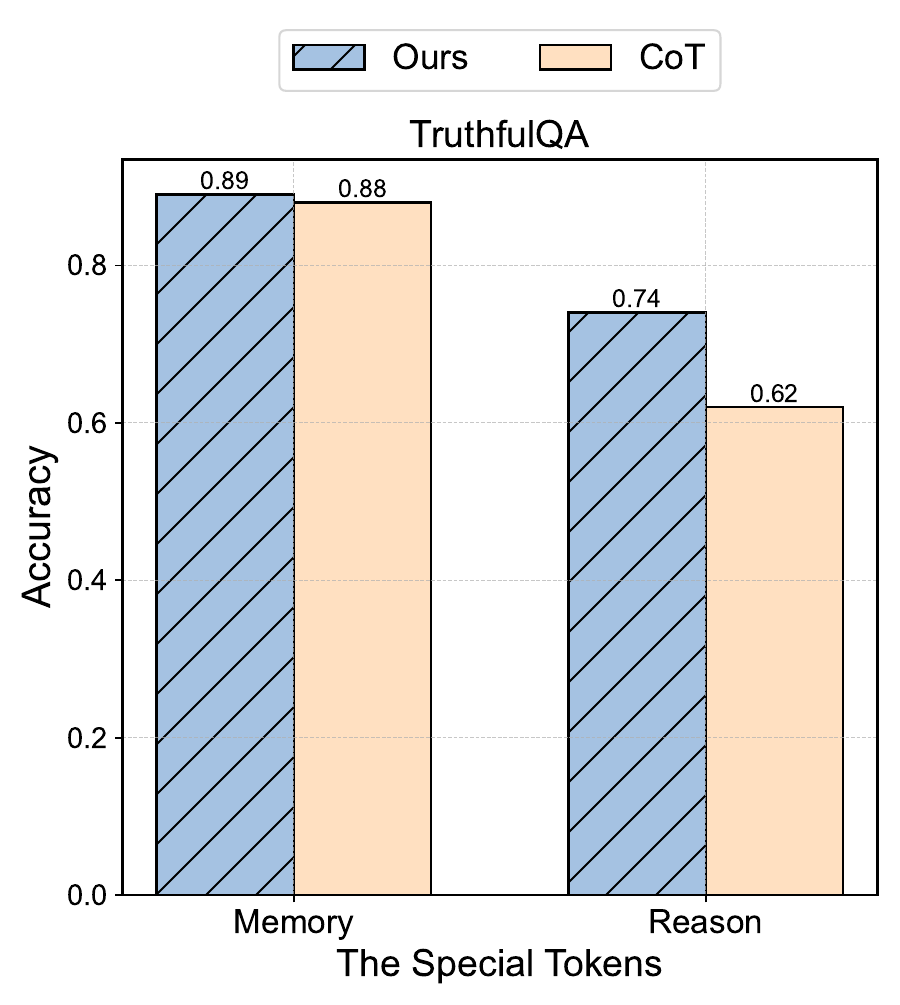}
    \end{minipage}
    \hfill
    \begin{minipage}[t]{0.31\textwidth}
        \centering
        \includegraphics[width=\textwidth]{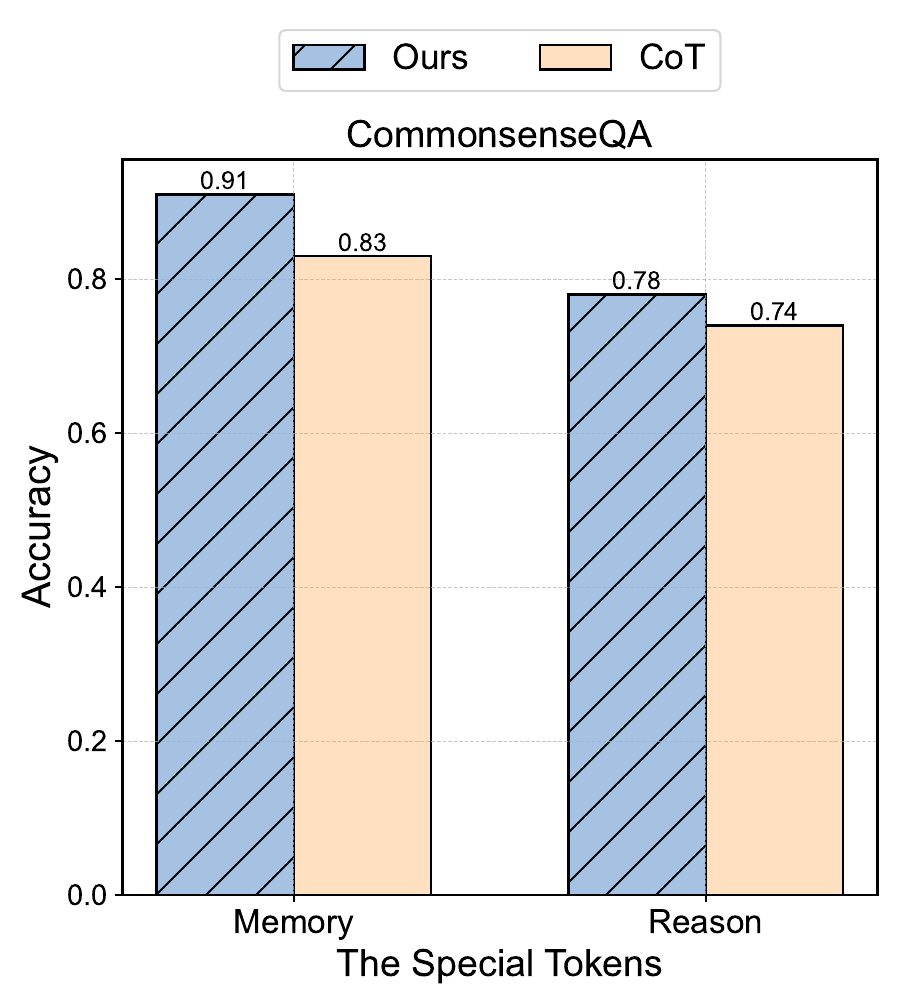}
    \end{minipage}
    \caption{Decoupling Result Comparison Between Our Algorithm and One-Shot CoT prompting on all datasets and both on LLaMA-3.1-8B, Accuracy stands for the decoupling performance of <memory> and <reason>.}
    \label{fig:case_study}
\end{figure*}


\begin{table}[ht]
    \centering
    \caption{Performance Comparison between One-shot CoT and our algorithm on LLaMA-3.1-8B.} 
    \label{table:method_comparison}
    \setlength{\belowcaptionskip}{-0.2cm}
    {
    \setlength{\tabcolsep}{20.0pt}  
    \small
    \begin{threeparttable}
    \begin{tabular}{@{}lc@{}}
        \toprule
         \textbf{Method} & \textbf{Accuracy (\% $\uparrow$)} \\
         \midrule
         \rowcolor[RGB]{234, 238, 234} \multicolumn{2}{c}{\textbf{StrategyQA}} \\
         One-shot CoT & 58.0 \\
         Ours & \textbf{78.0} \\
         \midrule
         \rowcolor[RGB]{234, 238, 234}
         \multicolumn{2}{c}{\textbf{CommonsenseQA}} \\
         One-shot CoT & 56.0 \\
         Ours & \textbf{82.3} \\
         \midrule
         \rowcolor[RGB]{234, 238, 234}
         \multicolumn{2}{c}{\textbf{TruthfulQA}} \\
         One-shot CoT & 54.0 \\
         Ours & \textbf{86.6} \\
        \bottomrule
    \end{tabular}
    \end{threeparttable}
    }
\vspace{-0.1em}
\end{table}

In this CoT approach, directive prompting with \( P_{\text{directive}} \) explicitly instructs the model to distinguish memory information and reasoning steps. The one-shot example \( E_{\text{1-shot}} \) provides a structured format, demonstrating how memory (e.g., \( M_1, M_2, \ldots, M_m \)) and reasoning parts (e.g., \( R_1, R_2, \ldots, R_n \)) can be organized separately in answer generation (e.g., \(M_1, R_1, M_2, R_2, \ldots, M_m, R_n\)
). This structure guides the model to produce answer and inference steps annotated as either memory \( M_m \) or reasoning \( R_n \), enhancing interpretability by separating factual knowledge and reasoning processes.

From Figure~\ref{fig:case_study} and Table~\ref{table:method_comparison}, on the StrategyQA dataset, our method achieves an accuracy of 78.0\% on LLaMA-3.1-8B, outperforming the One-shot CoT baseline by \textbf{20\%}, our approach achieves higher accuracy in decoupling memory (94\% vs. 93\%) and reasoning (71\% vs. 67\%), demonstrating effective decoupling between these two components in multi-steps inference. On the CommonsenseQA dataset, our method achieves an accuracy of 82.3\% on LLaMA-3.1-8B, exceeding the One-shot CoT baseline by \textbf{26.3\%}. The results highlight that our approach consistently outperforms the baseline in decoupling memory (91\% vs. 83\%) and reasoning (78\% vs. 74\%), demonstrating robust performance in commonsense inference tasks. On the TruthfulQA dataset, our method achieves an accuracy of 86.6\% on LLaMA-3.1-8B, surpassing the One-shot CoT baseline by \textbf{32.6\%}. The results further illustrate that our approach achieves superior accuracy in decoupling memory (89\% vs. 88\%) and reasoning (74\% vs. 62\%), highlighting its effectiveness in factual reasoning.
Additionally, Table~\ref{table:ratio_comparison} shows that both LLaMA-3.1-8B and LLaMA-2-7B maintain consistent distributions of memory and reasoning across all datasets. This reflects the stability and generalizability of our decoupling mechanism, ensuring its applicability to diverse inference tasks.

\begin{table}[ht]
    \centering
    \caption{[Memory:Reason] Ratio Across Different Models and Datasets.}
    \label{table:ratio_comparison}
    \setlength{\belowcaptionskip}{-0.2cm}
    {
    \setlength{\tabcolsep}{36.0pt} 
    \small
    \begin{threeparttable}
    \begin{tabular}{@{}lc@{}}
        \toprule
        \textbf{Method} & \textbf{Ratio} \\
        \midrule
        \rowcolor[RGB]{234, 238, 234} \multicolumn{2}{c}{\textbf{StrategyQA}} \\
        LLaMA-2-7B & 4:5 \\
        LLaMA-3.1-8B & 1:1 \\
        \midrule
        \rowcolor[RGB]{234, 238, 234} \multicolumn{2}{c}{\textbf{CommonsenseQA}} \\
        LLaMA-2-7B & 1:5 \\
        LLaMA-3.1-8B & 1:5 \\
        \midrule
        \rowcolor[RGB]{234, 238, 234} \multicolumn{2}{c}{\textbf{TruthfulQA}} \\
        LLaMA-2-7B & 3:7 \\
        LLaMA-3.1-8B & 1:2 \\
        \bottomrule
    \end{tabular}
    \end{threeparttable}
    }
    \vspace{-1.0em}
\end{table}
\subsubsection{Attention Analysis}
\label{case_study:attention_analysis}
\begin{figure*}[!h]
    \centering
    \begin{minipage}{0.48\textwidth}
        \centering
        \includegraphics[width=\linewidth]{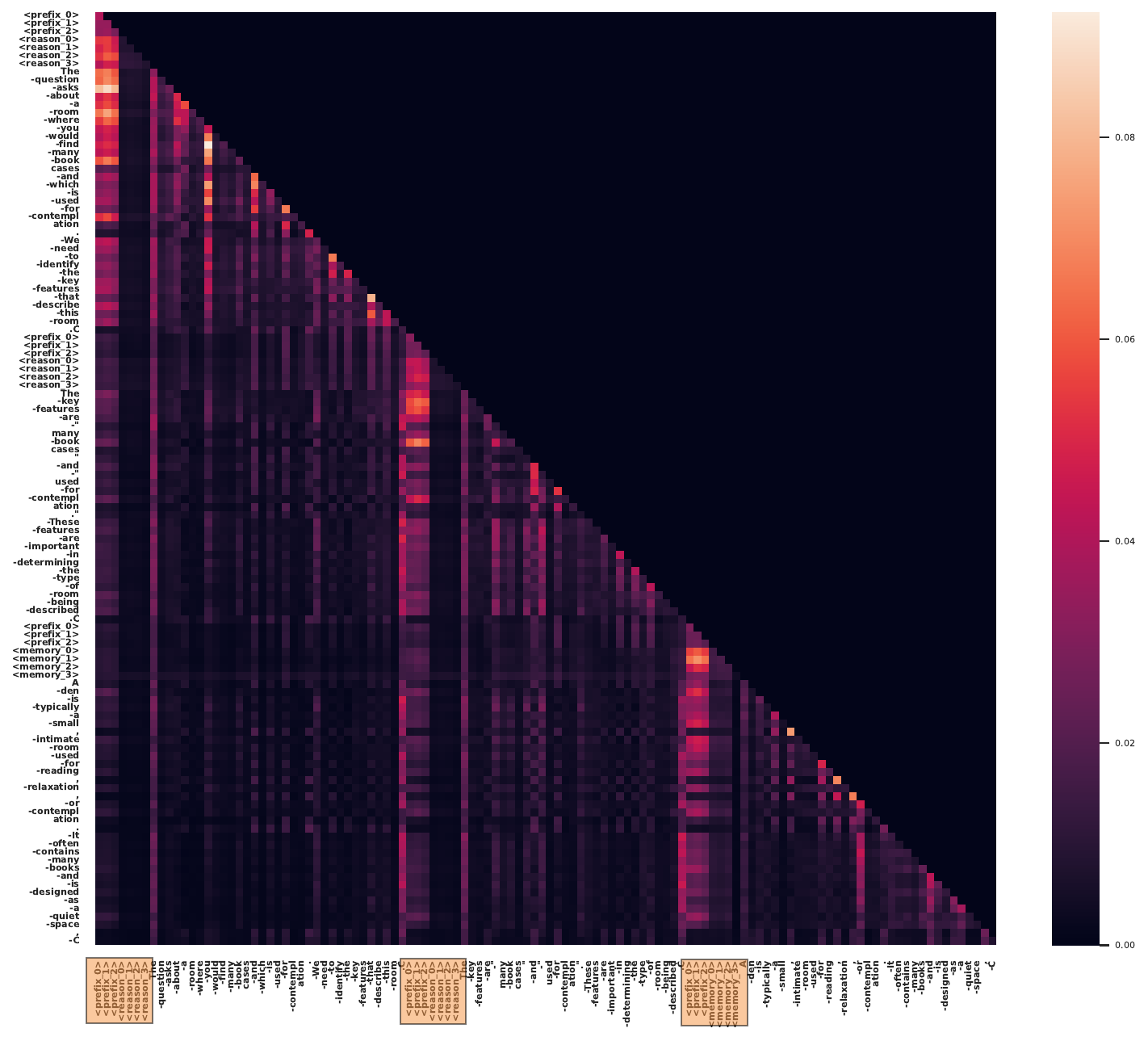}
        \label{fig:LLaMA3-8b}
    \end{minipage}
    \hfill
    \begin{minipage}{0.48\textwidth}
        \centering
        \includegraphics[width=\linewidth]{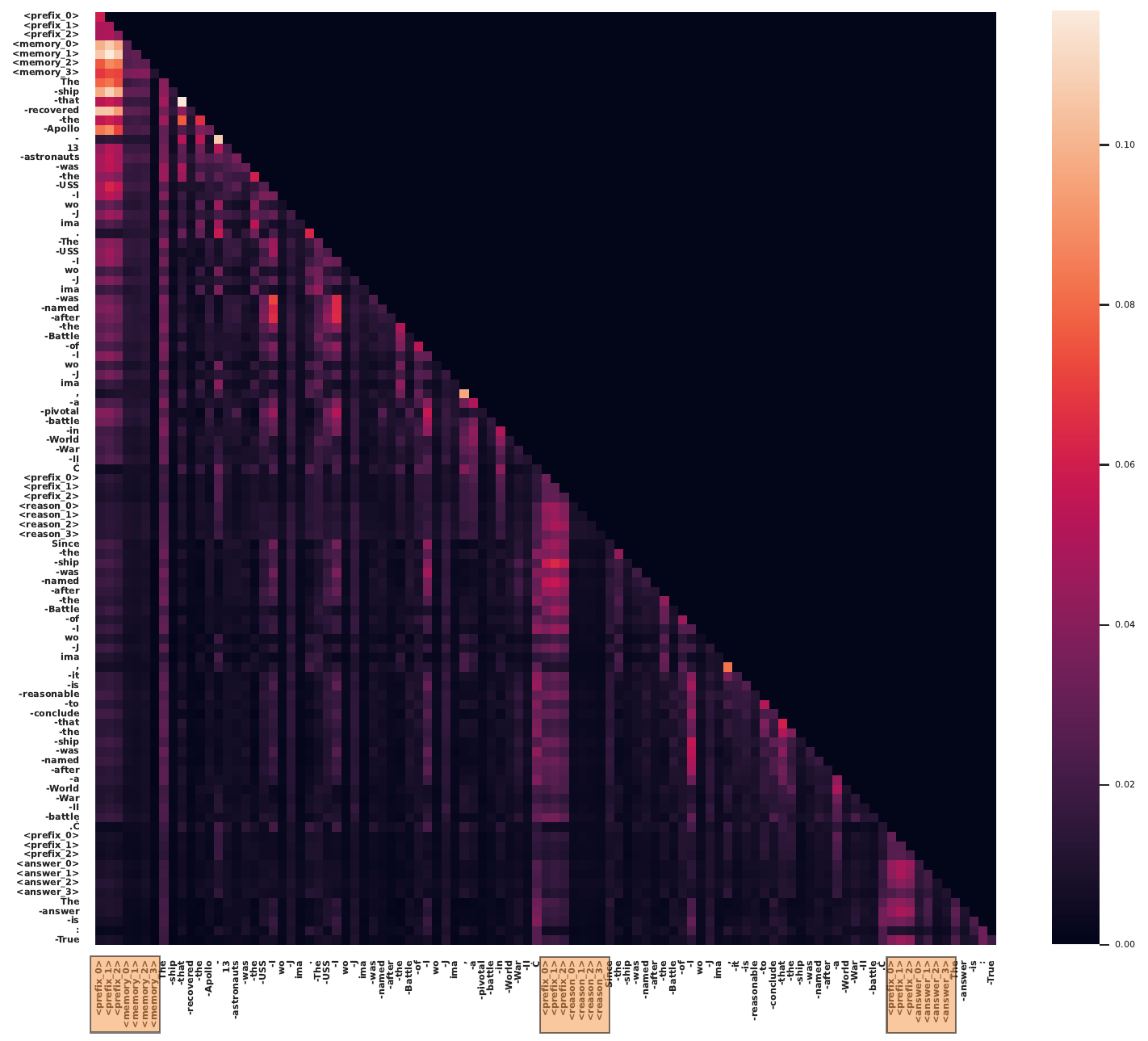}
        \label{fig: LLaMA-2-7b}
    \end{minipage}
    \hfill
    \vspace{-1.2em}
     \caption{Two test examples' attention Heatmap generated by LLaMA-3.1-8B enhanced with our algorithm in the same attention head. The highlighted parts are these special tokens.}
     \label{attention}
\end{figure*}

In the case study~\ref{case_study:sample_analysis}, we have found that the {$\langle \text{reason} \rangle$} and {$\langle \text{memory} \rangle$} do an important job in our LLM's Inference. Although using raw attention weights to interpret token importance can be somewhat controversial, attention patterns still provide valuable insights about how transformers operate~\cite{abnar2020quantifying}. This heatmap, as Figure \ref{attention} shows that the model focuses intensely on specialized tokens throughout the inference. These tokens received higher attention weights than regular tokens, suggesting they play a more significant role in leading knowledge and reasoning content generation. This observation aligns with the main findings presented in the previous case study~\ref{case_study:sample_analysis}.

We input two sentences (can be found in Appendix~\ref{app:case_study_more_attention_maps} in Figure~\ref{fig:app:prompt_attention_map}) into our fine-tuned LLaMA-3.1-8B model, getting a large attention heatmap. We then segmented two entire attention maps according to the steps by model inference, producing the two smaller maps above as Figure~\ref{attention}. Other samples can be found in the Appendix~\ref{app:case_study_more_attention_maps}. by the observation, it indicate that the model places greater emphasis on this content, which indirectly demonstrates the effectiveness of our algorithm.



\begin{table}[ht]
    \centering
    \caption{Error Type Proportion between Memory and Reason on LLaMA-3.1-8B across all the datasets.} 
    \label{table:error_analysis}
    \setlength{\belowcaptionskip}{-0.2cm}
    {
    \setlength{\tabcolsep}{30.0pt}  
    \small
    \begin{threeparttable}
    \begin{tabular}{@{}lc@{}}
        \toprule
         \textbf{Error Type} & \textbf{Proportion} \\
         \midrule
         \rowcolor[RGB]{234, 238, 234} \multicolumn{2}{c}{\textbf{StrategyQA}} \\
         Memory & 1.7 \\
         Reason & 98.3 \\
         \midrule
         \rowcolor[RGB]{234, 238, 234}
         \multicolumn{2}{c}{\textbf{CommonsenseQA}} \\
         Memory & 21.6 \\
         Reason & 78.4 \\
         \midrule
         \rowcolor[RGB]{234, 238, 234}
         \multicolumn{2}{c}{\textbf{TruthfulQA}} \\
         Memory & 21.1 \\
         Reason & 78.9 \\
        \bottomrule
    \end{tabular}
    \end{threeparttable}
    }
\end{table}

\subsubsection{Error Analysis}
\label{case_study:error_analysis}
\begin{figure*}[!th]
\centering
\includegraphics[width=\linewidth]{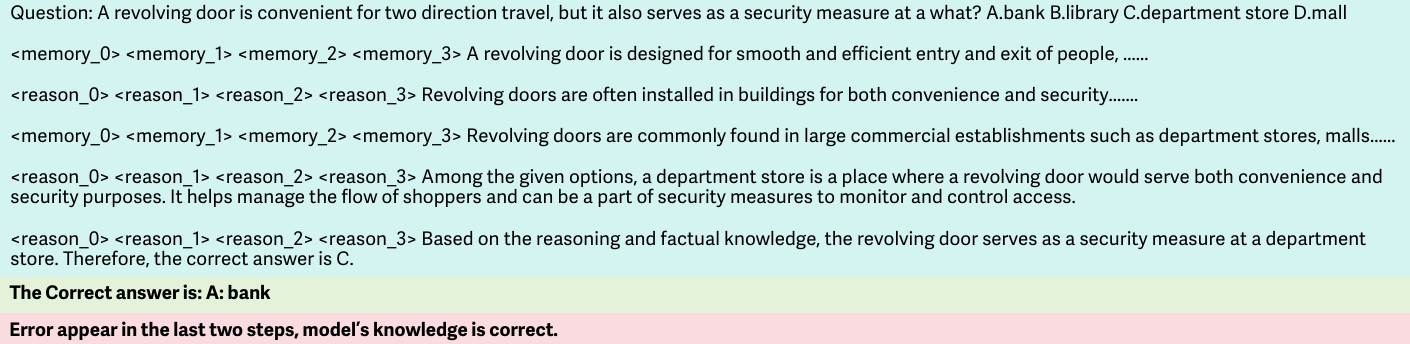}
\caption{\textbf{Incorrect Sample Showing:} The green sections represent the questions, the steps of model inference, and the incorrect answers; the yellow areas indicate the correct answers, and the red highlights the causes of the errors.}
\label{fig:input}
\vspace{-1em}
\end{figure*}

We analyzed all incorrect results generated by our fine-tuned LLaMA-3.1-8B model to identify whether the errors originated from memory or reasoning issues, utilizing GPT-4o to categorize the source of each error across StrategyQA, CommonsenseQA, and TruthfulQA benchmarks. As shown in Table~\ref{table:error_analysis}, 98.3\% of the errors in StrategyQA were attributed to reasoning, with only 1.7\% due to memory issues, indicating reasoning as the dominant challenge. Similarly, in CommonsenseQA, 78.4\% of errors stemmed from reasoning, while 21.6\% were caused by memory failures; in TruthfulQA, the trend persisted, with 78.9\% of errors linked to reasoning and 21.1\% to memory. These results demonstrate that reasoning-related errors consistently account for over 75\% of total mistakes across benchmarks, underscoring that while the model successfully utilizes knowledge, it requires significant improvements in reasoning capabilities, pointing to an important direction for future research.

For Example, the question \ref{fig:input} above emphasizes the role of the revolving door as a security measure, so the correct answer should be somewhat unexpected. Options B, C, and D all represent typical uses of revolving doors for managing two-way traffic flow. Only in banks does a revolving door serve as a security measure. The correct answer is likely A. bank, as banks use revolving doors not only for easy access but also as a security measure to control entry and exit. The model’s knowledge is accurate, but it missed this nuance during reasoning steps.

\section{Related Work}

\paragraph{Parametric Memory in LLMs.} During pre-training, large language models capture a large amount of knowledge in the models' parameters, known as parametric memory. Previous research extensively explores the mechanism of inference with parametric memory; they observe that models can well adopt memory for simple tasks but struggle with complex inference, \textit{e.g.,} multi-hop inference \citep{li2024understandingpatchingcompositionalreasoning,yang2024largelanguagemodelslatently,wang2024grokkedtransformersimplicitreasoners, jin2025exploring, han2024token, wang2025larger}. Others reveal the challenges in the leverage of parametric knowledge, particularly when dealing with long-tail facts (facts associated with less common entities) or when the knowledge is rare \citep{wang2023survey,allen2023physics,cheng2024understandinginterplayparametriccontextual, jin2025massive}. These studies primarily focus on the analysis of model behavior. While valuable, they do not address how to better elicit the parametric knowledge for inference. In this work, we explore how to boost LLMs' leverage of their parametric knowledge for complex inference.

\paragraph{Reasoning with LLMs.}

Recent research on enhancing LLMs' inference capabilities can be broadly categorized into prompt-based and tuning-based approaches. 
Prompt-based methods strategically guide reasoning processes. Chain-of-Thought (CoT) prompting~\cite{wei2022chain} and its derivatives~\cite{zhao-etal-2024-enhancing-zero, zhou2023threadthoughtunravelingchaotic, chen2024boostingthoughtstrialanderrorproblem, hu2024chainofsymbolpromptingelicitsplanning,jin2024impact} decompose complex tasks into sequential steps, improving transparency and decision-making. Others like Tree-of-Thoughts (ToT)~\cite{NEURIPS2023_271db992} and Graph-of-Thoughts (GoT)~\cite{besta2024got} further utilize hierarchical and network-based inference to cover a larger search space. These strategies design a framework where LLMs can elicit the parametric memory relevant to the task. However, in these methods, the models might not know when to reason or use their memory.

Tuning-based methods introduce trainable tokens for structured CoT steps, facilitating reasoning and utilization of memory \cite{wang2024guidinglanguagemodelreasoning,goyal2024thinkspeaktraininglanguage,colonhernandez2024languagemodelshintprompting,wang2025dump}.
Despite the effectiveness, these methods intertwine reasoning and memory usage, which may limit the full potential of the models. In contrast, our approach aims to decouple memory and reasoning within the CoT process by introducing various special tokens, enabling the model to leverage its memory more effectively.

\section{Conclusion}
In this work, we proposed a novel inference framework for training LLMs to distinguish between reasoning and memory processes using two special tokens: $\langle \text{memory} \rangle$ for factual knowledge retrieval and $\langle \text{reason} \rangle$ for logical reasoning. This structured input disentangles these processes, enhancing interpretability and improving performance on complex reasoning tasks. By maintaining a clear boundary between memory and reasoning during training, the model generalizes better to queries that combine factual knowledge with multi-step reasoning. This approach not only ensures more accurate answers but also produces interpretable, step-by-step reasoning outputs, crucial for transparency and accountability in complex reasoning.

\section{Limitation}
The proposed decomposition framework provides a promising method to disentangle memory recall and reasoning in large language models, enhancing interpretability and modularity. However, it has limitations that offer opportunities for improvement. One challenge is its reliance on the quality and breadth of training data for memory recall, which may lead to incomplete retrieval in underrepresented domains. This issue, common in machine learning, can be mitigated through dynamic updates or integration with external knowledge bases. The use of special tokens like $\langle \text{memory} \rangle$ and $\langle \text{reason} \rangle$ simplifies distinguishing between tasks but adds complexity to tokenization, requiring task-specific tuning for different architectures or languages. Nonetheless, this token-based design enhances transparency, offsetting the added complexity. The framework also struggles with tasks requiring deeply nested or multi-hop reasoning, as these steps may not neatly separate into recall and reasoning phases. Further refinement is needed to better handle complex reasoning chains, though the framework performs robustly in standard scenarios. Additionally, the retrieval-based approach introduces computational overhead, which may limit real-time applicability. However, the trade-off for interpretability and error traceability is valuable for use cases where transparency is critical, making this framework a significant step forward for addressing reasoning and memory in LLMs

\bibliography{custom}

\appendix
\newpage
\centerline{\maketitle{\textbf{SUMMARY OF THE APPENDIX}}}

This appendix contains additional details for the \textbf{\textit{``Disentangling Memory and Reasoning Ability
in Large Language Models''}}. The appendix is organized as follows:

\begin{itemize}
    \item \S\ref{app:data} \textbf{Data Generation}
    \begin{itemize}
        \item \ref{app:data:implement}~Implement Details
        \item \ref{app:data:example}~Example
    \end{itemize}

    \item \S\ref{appendix:preliminary_experiment} \textbf{Preliminary Study}
    \begin{itemize}
        \item \ref{appendix:preliminary_experiment:experiment}~Experiment
        \item\ref{appendix:preliminary_experiment:example}~Example
    \end{itemize}
    
    \item \S\ref{app:metric} \textbf{Experiment Details}
    \begin{itemize}
        \item \ref{app:metric:dataset}~Dataset
        \item \ref{app:metric:evaluation_metric}~Evaluation Metric
    \end{itemize}

    \item \S\ref{app:train} \textbf{Training Details}
    \begin{itemize}
        \item \ref{app:train:training_configuration}~Training Configuration
        \item \ref{app:train:training Process}~Training Process
        \item \ref{app:training_details:example}~Example
    \end{itemize}
    
    \item \S\ref{appendix:case_study} \textbf{Case Study}
    \begin{itemize}
        \item \ref{appendix:case_study:Sample Analysis}~Sample Analysis
        \item\ref{app:case_study:error_analysis}~Error Analysis
        \item\ref{app:case_study_more_attention_maps}~More Attention Maps

        \item\ref{app:case_study_more_analysis_samples}~More Analysis Samples
    \end{itemize}
     \item \S\ref{app:future_Work_and_limitation} \textbf{Future Work and Limitation}
     \begin{itemize}
        \item \ref{app:future_Work_and_limitation:future_work}~Future Work
        \item \ref{app:future_Work_and_limitation:limitation}~Limitation
    \end{itemize}

\end{itemize}

\section{Data Generation}
\label{app:data}
\subsection{Implement Details}
\label{app:data:implement}
We developed an LLM-based data generation framework based on GPT-4o to generate high-quality training data for decoupling memory and reasoning steps. This framework includes two LLMs: an inference LLM and a knowledge LLM. The inference LLM is responsible for generating Chain-of-Thought (CoT) inference processes, decoupling memory and reasoning, and then further assigning labels to each sub-step by marking those requiring factual knowledge as [memory] and those requiring reasoning as [reason]. The prompt of the knowledge agent is shown in Figure~\ref{fig:prompt_in_knowledge_llm}. The Knowledge LLM retrieves the necessary knowledge for [memory] steps based on questions provided by the inference LLM. We use these two LLMs to ensure the independence of memory and reasoning within the CoT, providing high-quality data for subsequent training. Figure~\ref{fig:prompt_in_inference_LLM} is the prompt configuration for inference LLM. The \textit{questions} corresponds to the \textit{Knowledge base} content in the inference LLM and is used to supply accurate factual information for steps labeled as <memory>.
\begin{figure}[ht]
    \centering
    \begin{tcolorbox}[
        title=\texttt{Prompt in Knowledge LLM},
        width=0.47\textwidth 
    ]
    \begin{flushleft}
        Factual knowledge is information that aligns with objective reality and can be verified through evidence or observation, such as scientific facts or historical events.\\
        \vspace{1em}
        Please provide factual knowledge for the below question set:\\
        <Questions>\\
        \{\textbf{questions}\}\\
        <Questions>\\
        \vspace{1em}

        You should return a dictionary in JSON format; for each element in the dictionary, the key is each question in <Questions>, and the value is the factual knowledge of each question in <Questions>.\\
        \vspace{1em}
        Your answer format should strictly be in the following steps:\\
        ```json\\
        \{\\
             \qquad "Question 1": "The factual knowledge of question 1",\\
        \qquad ........\\
        \}
        ```
    \end{flushleft}
    \end{tcolorbox}
    \caption{Prompt in Knowledge LLM to activate the inner knowledge}
    \label{fig:prompt_in_knowledge_llm}
\end{figure}
 The \textit{Step name} refers to the specific name of each step in the Chain of Thought (CoT) process. The \textit{Requirement} labels whether each step pertains to <memory> or <reason>. The \textit{Knowledge based} is used to provide questions related to factual knowledge in <memory> steps, while the \textit{Content} focuses on designing to outline the reasoning process for <reason> steps. This structure facilitates a clear distinction between memory retrieval and reasoning tasks, enhancing the model's capability to execute complex sequences in a zero-shot environment.
\begin{figure*}[ht]
    \centering
    \begin{tcolorbox}[
        title=\texttt{Prompt in Inference LLM},
        width=\textwidth 
    ]
    \begin{flushleft}
        Here is the question:\\ 
        <Question>\\
        \{\textbf{question}\}\\
        <Question>\\
        \vspace{1em}
        Here is the correct answer:\\
        <Correct Answer>\\
        \{\textbf{answer}\}\\
        <Correct Answer>\\
        \vspace{1em}
        Factual knowledge is information that aligns with objective reality and can be verified through evidence or observation, such as scientific facts or historical events.\\
        \vspace{1em}
        Provide a reasoning plan for the above question to get the correct answer; each step in your reasoning plan must adhere strictly to the following format:\\
        \vspace{1em}
        \textcolor{darkblue}{*Step name*: }\\
        \# Put the name of the step here.\par
        \vspace{1em}

        \textcolor{darkblue}{**Requirement**: }\\
        \# If this step needs reasoning, return "[reason]" as a label; if this step needs factual knowledge, return "[memory]" as a label.\\
        \vspace{1em}

        \textcolor{darkblue}{**Knowledge based**: }\\
        \# Only if this step needs factual knowledge, put a query in question sentences about this factual knowledge for retrieval.\\
        \vspace{1em}

        \textcolor{darkblue}{**Content**: }\\
        \# If this step is about reasoning, please provide your reasoning thinking; if this step needs factual knowledge, please provide factual knowledge.\\
    \end{flushleft}
    \end{tcolorbox}
    \caption{Prompt in Inference LLM}
    \label{fig:prompt_in_inference_LLM}
\end{figure*}

\subsection{Example}
\label{app:data:example}

In this study, we leverage both the inference LLM and knowledge LLM based on GPT-4o to generate a dataset. Specifically, we use the StrategyQA dataset as the source for our generation. The StrategyQA dataset is a question-answering dataset designed to evaluate models' multi-hop reasoning abilities. It includes questions that require strategic thinking and often demand more than one piece of information to answer correctly.\par
Figure~\ref{fig:enhanced_dataset} shows enhanced data generated by our data generation agent from the StrategyQA dataset. The image shows an example question about the relationship between Mixed Martial Arts and the origins of Roman Colosseum games. The answering process is broken down into several steps, each labeled as either [Memory] or [Reason] to indicate the type of step. This approach helps differentiate between pure knowledge retrieval and logical reasoning steps, providing more granular training data for models to improve their accuracy and interpretability in answering complex questions.

\begin{figure*}[h]
    \centering
    \begin{tcolorbox}[
        title=\texttt{LLM enhanced dataset(StrategyQA)},
        width=\textwidth 
    ]
    \begin{flushleft}
        \textcolor{darkred}{\textbf{Question:}} Is Mixed martial arts totally original from Roman Colosseum games?  \\
        \textcolor{darkblue}{\textbf{Solution Step:}}
        \begin{enumerate}
            \item "[Memory]: Mixed Martial Arts (MMA) is a full-contact combat sport that allows a wide variety of fighting techniques from different martial arts traditions. It permits both striking and grappling, both standing and on the ground, using techniques from disciplines such as boxing, wrestling, Muay Thai, karate, and judo."
            \item "[Memory]: The Roman Colosseum games, also known as gladiatorial games, were violent contests where gladiators fought against each other, condemned criminals, or wild animals. These events were held in large amphitheaters like the Colosseum in Rome and were a form of public spectacle and entertainment in ancient Rome."
            \item "[Memory]: Modern MMA is characterized by regulated rules, weight classes, and a combination of various martial arts disciplines. It is officiated with rules to ensure the safety of participants, and fights occur in a controlled environment, often inside a cage."
            \item "[reason]: MMA and the Roman Colosseum games share the concept of hand-to-hand combat but differ significantly in purpose, structure, and regulation. While MMA is a sport with rules designed for competition and fighter safety, the Roman games were more about public spectacle and entertainment without much emphasis on fairness or safety. The combat in Roman games was often deadly and executed for the spectators' pleasure."
            \item "[reason]: MMA is not totally original from the Roman Colosseum games. Although both involve unarmed combat, MMA is a modern sporting discipline that synthesizes traditional martial arts into a competitive and regulated environment. The Roman games served as a historical precedent for public combat events but lacked the structured and safety-oriented approach of MMA. Therefore, while there may be a historical inspiration, MMA's development as a technical and regulated sport makes it distinct and not directly derived from the Roman games."
        \end{enumerate}
        \textcolor{darkpurple}{\textbf{Answer:} False} 
    \end{flushleft}
    \end{tcolorbox}
    \caption{StrategyQA dataset example(enhanced by our algorithm)}
    \label{fig:enhanced_dataset}
\end{figure*}

\section{Preliminary}
\label{appendix:preliminary_experiment}

\subsection{Experiment}
\label{appendix:preliminary_experiment:experiment}
\begin{figure}[h]
    \centering
    \includegraphics[width=0.4\textwidth]{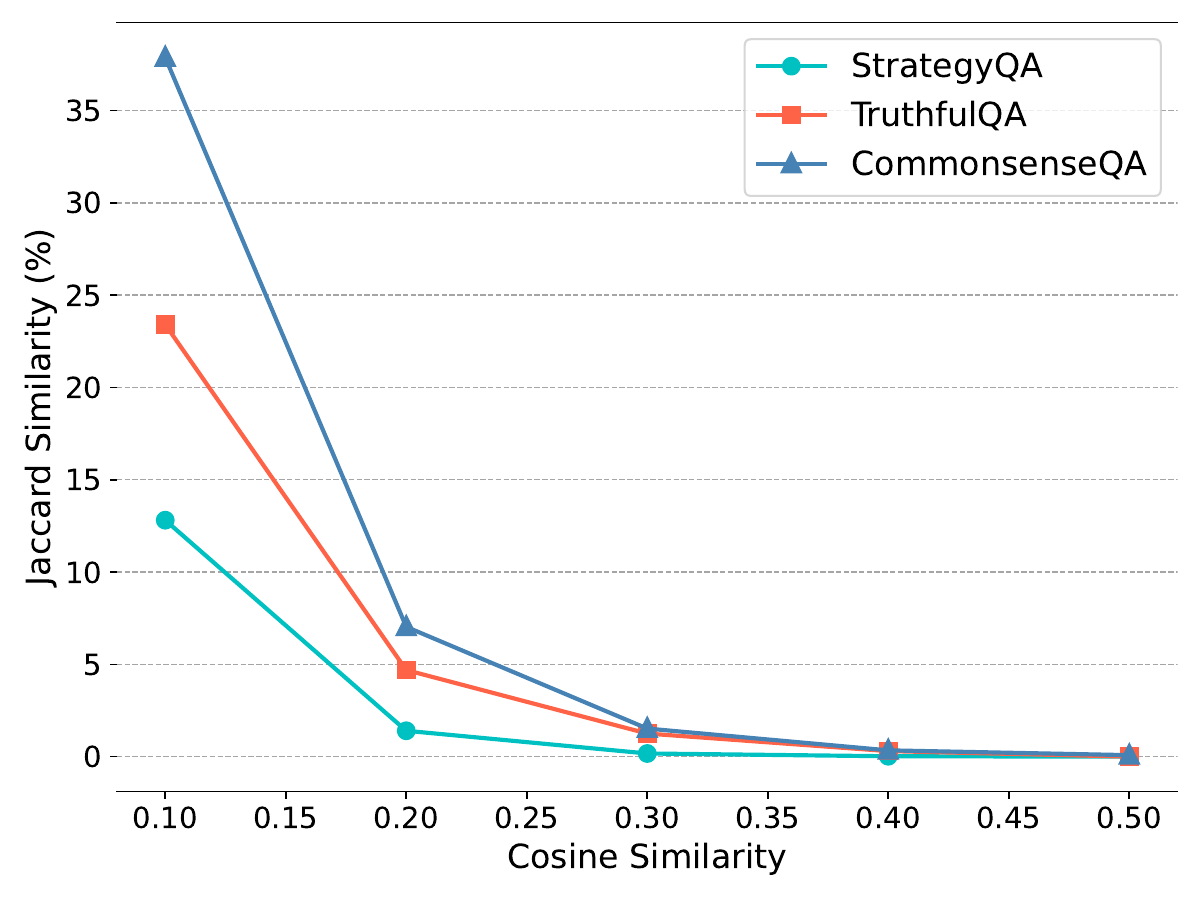}
    \captionof{figure}{Jaccard Similarity for Generated and Training Data}
    \label{preliminary}
\end{figure}

In a preliminary experiment, we analyzed the training and test sets of StrategyQA, TruthfulQA, and CommonsenseQA to evaluate the overlap in knowledge between them. This assessment was crucial to ensure that \textit{our model's performance improvement was due to our advanced algorithm, rather than simply distilling knowledge from GPT-4o}. For our synthetic training set, we extracted sentences following each $\langle \text{memory} \rangle$ token to create a reference set. We then prompted our fine-tuned LLaMA3.1-8B model to generate outputs using the test set, collecting sentences following the $\langle \text{memory} \rangle$ tokens in these outputs to form a separate set. Our validation method involves setting a threshold on cosine similarity and assessing Jaccard similarity based on this threshold. Specifically, as illustrated in Figure~\ref{preliminary}, we define two knowledge after $\langle \text{memory} \rangle$ tokens as overlapping if the cosine similarity of their embeddings exceeds 0.2. Based on this criterion, the Jaccard similarity for all datasets are smaller than 10\%, which is a low value indicating a low degree of overlap and demonstrates that our model’s performance is not merely the result of knowledge distillation.

\subsection{Example}
\label{appendix:preliminary_experiment:example}
A value greater than or equal to 0.2 indicates that the two contents are very unrelated like example~\ref{fig:cosine_similarity_example}. 
\begin{figure}[h]
    \begin{tcolorbox}[
        title=\footnotesize\texttt{Cosine Similarity of a sample in Testset and Trainset},
        fontupper=\footnotesize,
        colback=white,
        colframe=black,
        boxrule=0.5pt,
        top=0.5em,
        bottom=0.5em,
        left=0.5em,
        right=0.5em
    ]
    \begin{flushleft}
        \textcolor{darkred}{\textbf{Testset:}} The question asks about a type of store that would have a lot of sports equipment. This requires understanding what type of store would typically sell a variety of sports-related items.\\
        \vspace{0.5em}
        \textcolor{darkblue}{\textbf{Trainset:}} Sainsbury's and Tesco are both publicly traded companies. As of the latest available data, Tesco's market capitalization is significantly larger than that of Sainsbury's. For Sainsbury's to acquire Tesco, it would require extensive financial resources or backing, potentially involving significant borrowing, asset sales, or equity raising.\\
        \vspace{0.5em}
        \textcolor{darkpurple}{\textbf{Cosine Similarity: }}0.2
    \end{flushleft}
    \end{tcolorbox}
    \caption{A sample in Testset and Trainset}
    \label{fig:cosine_similarity_example}
\end{figure}

\section{Experiment Details}
\label{app:metric}
\subsection{Dataset}
\label{app:metric:dataset}
\paragraph{StrategyQA~\cite{geva2021strategyqa}}
StrategyQA is a challenging question-answering benchmark that focuses on implicit, multi-step reasoning. Unlike conventional multi-hop datasets where questions explicitly outline the steps needed to reach an answer, StrategyQA requires models to infer these reasoning steps. Each question in StrategyQA is crafted to be implicit and short, with Boolean ("Yes" or "No") answers, requiring logical deductions based on general knowledge. For example, answering a question like “Did Aristotle use a laptop?” involves reasoning about the historical timeline of both Aristotle’s life and the invention of laptops. The StrategyQA dataset includes a total of 2,780 verified questions. The training set comprises 1,600 questions which are used for fine-tuning, and the validation (test) set contains 690 questions, which are used for the validation of baselines and our method in our experiment.
\paragraph{CommonsenseQA~\cite{talmor-etal-2019-commonsenseqa}}
CommonsenseQA is a multiple-choice dataset with 12,247 questions aimed at testing AI on commonsense reasoning using the ConceptNet knowledge graph. Each question has one correct answer and four distractors, requiring models to understand relations like causality and spatial proximity. While humans achieve 88.9\% accuracy, advanced models like BERT-Large reach only 55.9\%, underscoring the challenge of commonsense inference in AI. The training set comprises 9,740 questions, the validation set contains 1,220 questions, which are used for fine-tuning, and the test set includes 1,140 questions, which are used for the validation of baselines and our method in our experiment.
\paragraph{TruthfulQA~\cite{lin2022truthfulqa}}
TruthfulQA is a benchmark of 817 questions designed to test language models' truthfulness by prompting common misconceptions across topics like health and law. Models like GPT-3 and GPT-2 often generate false answers that mirror human misunderstandings, with larger models frequently performing worse (58\% truthfulness for GPT-3) compared to 94\% for humans. The benchmark reveals that scaling up model size alone does not enhance truthfulness, highlighting the need for targeted fine-tuning to reduce imitative falsehoods. In our experiments, we split the dataset into training and testing sets in an 8:2 ratio. Since the original dataset contained only single-choice questions with all answers marked as \textit{A}, we randomly shuffled the answer options for one question to ensure effective fine-tuning performance on the training set.

\subsection{Evaluation Metric}
\label{app:metric:evaluation_metric}
To mitigate the inherent output instability of LLMs in both CoT and Zero-shot settings, we found that conventional answer-matching techniques, such as regular expression-based methods, may not reliably capture the precise answers required. Consequently, we adopted GPT-4o-mini as an evaluation tool to compute the LLM performance across multiple datasets~\cite{cao2024guidedefenseg4ddynamic}. This approach enables a more nuanced assessment of LLM outputs, given the limitations of regular matching techniques under these settings. The detailed prompt used for evaluation is shown in Prompt~\ref{fig:gpt_4o_mini_eval} below.

\begin{figure}[ht]
    \centering
    \begin{tcolorbox}[
        title=\texttt{Prompt in GPT-4o-mini},
        label={prompt3},
        width=0.5\textwidth 
    ]
    \begin{flushleft}
        You should only return True if the user gives the correct answer or the content related to the correct answer, otherwise, you should return False.\\
        \vspace{1em}
        \#\# Question:\\
        <BEGIN QUESTION>\\
        \{\textbf{questions}\}\\
        <END QUESTION>\\
        \vspace{1em}

        \#\# Correct Answer:\\
        <BEGIN CORRECT ANSWER>\\
        \{\textbf{correct answer}\}\\
        <END CORRECT ANSWER>\\
        \vspace{1em}

        \#\# User Answer:\\
        <BEGIN USER ANSWER>\\
        \{\textbf{user answer}\}\\
        <END USER ANSWER>\\
        \vspace{1em}
        \# Judgement:\\
        \#\# True or False:\\
    \end{flushleft}
    \end{tcolorbox}
    \caption{Prompt in GPT-4o-mini for Evaluating CoT Reasoning}
    \label{fig:gpt_4o_mini_eval}
\end{figure}

\section{Training Details}
\label{app:train}

\subsection{Training Configuration}
\label{app:train:training_configuration}
All the experiments for fine-tuning are run on an NVIDIA RTX 6000 Ada Generation GPU. Our experiments found that the optimal configuration for learning out-of-vocabulary (OOV) tokens is with N\_PREFIX=3 and N\_SPECIAL=4. We generally use a learning rate of 2e-4 with --warmup\_steps 1000, --lr\_scheduler\_type "cosine", and --optim "adamw\_torch", along with gradient\_accumulation\_steps=16. Additionally, we employed int8 training to ensure that the model could be trained on a single GPU. Additionally, We provided detailed parameter configurations as below:
Here is the detailed training configuration:\\

\subsection{Training Process}
\label{app:train:training Process}
\begin{figure*}[t]
    \centering
    \begin{minipage}{0.45\textwidth}
        \centering
        \includegraphics[width=\linewidth]{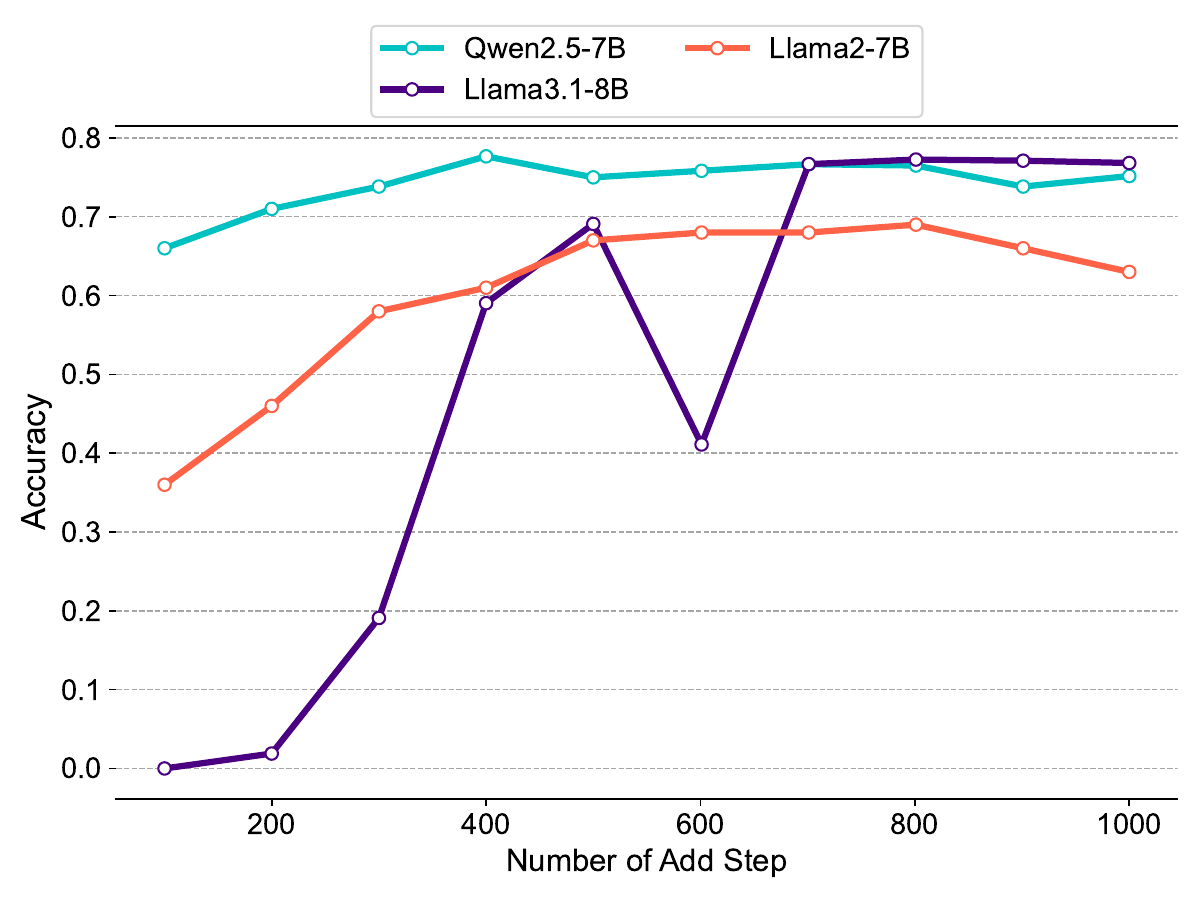}
        \vspace{-1em}
        \caption{Accuracy progression on the StrategyQA benchmark during training, with the horizontal axis representing the number of training steps.}
        \label{fig:image1}
    \end{minipage}%
    \vfill
    \begin{minipage}{0.45\textwidth}
        \centering
        \includegraphics[width=\linewidth]{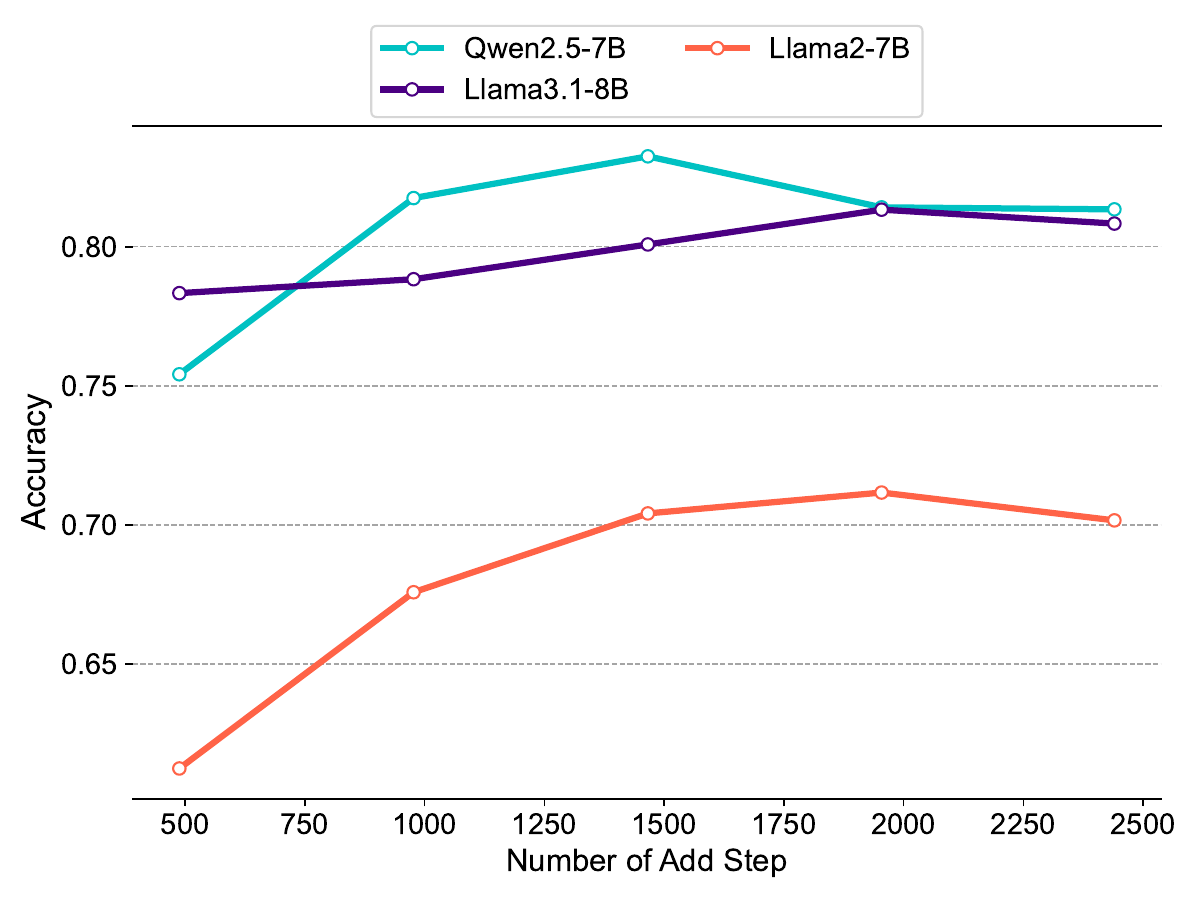}
        \caption{Accuracy progression on the CommonsenseQA benchmark during training. Qwen2.5-7B achieves the highest accuracy early on, followed by a stable plateau. }
        \label{fig:image2}
    \end{minipage}%
    \hspace{0.01\textwidth} 
    \begin{minipage}{0.45\textwidth}
        \centering
        \includegraphics[width=\linewidth]{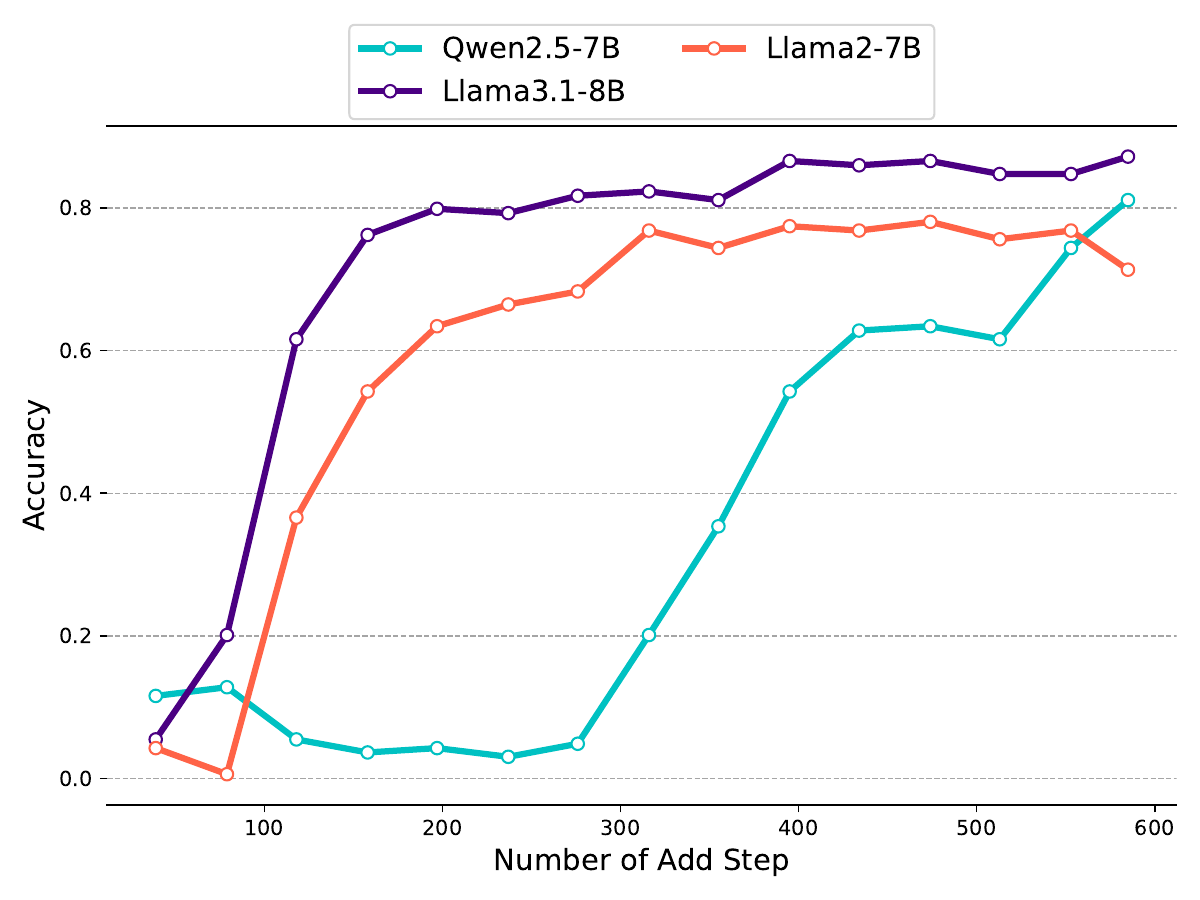}
        \caption{Accuracy progression on the TruthfulQA benchmark during training. Llama3.1-8B outperforms other models, showing rapid early improvement and reaching the highest accuracy.}
        \label{fig:image3}
    \end{minipage}
    \caption{The Training Process for LLaMA3.1-8B on three Datasets: for StratgyQA, we usually need ten epochs to train and five epochs to model to converge in CommonsenseQA. For TruthfulQA, 15 epochs or more may be better.}
    \label{fig:train}
\end{figure*}

We monitored the training process of our method across all models and datasets, recording test set accuracy changes every 10 steps, as illustrated in Figure~\ref{fig:train}.

\subsection{Example}
\label{app:training_details:example}
\begin{figure*}[ht]
    \centering
    \includegraphics[width=\textwidth]{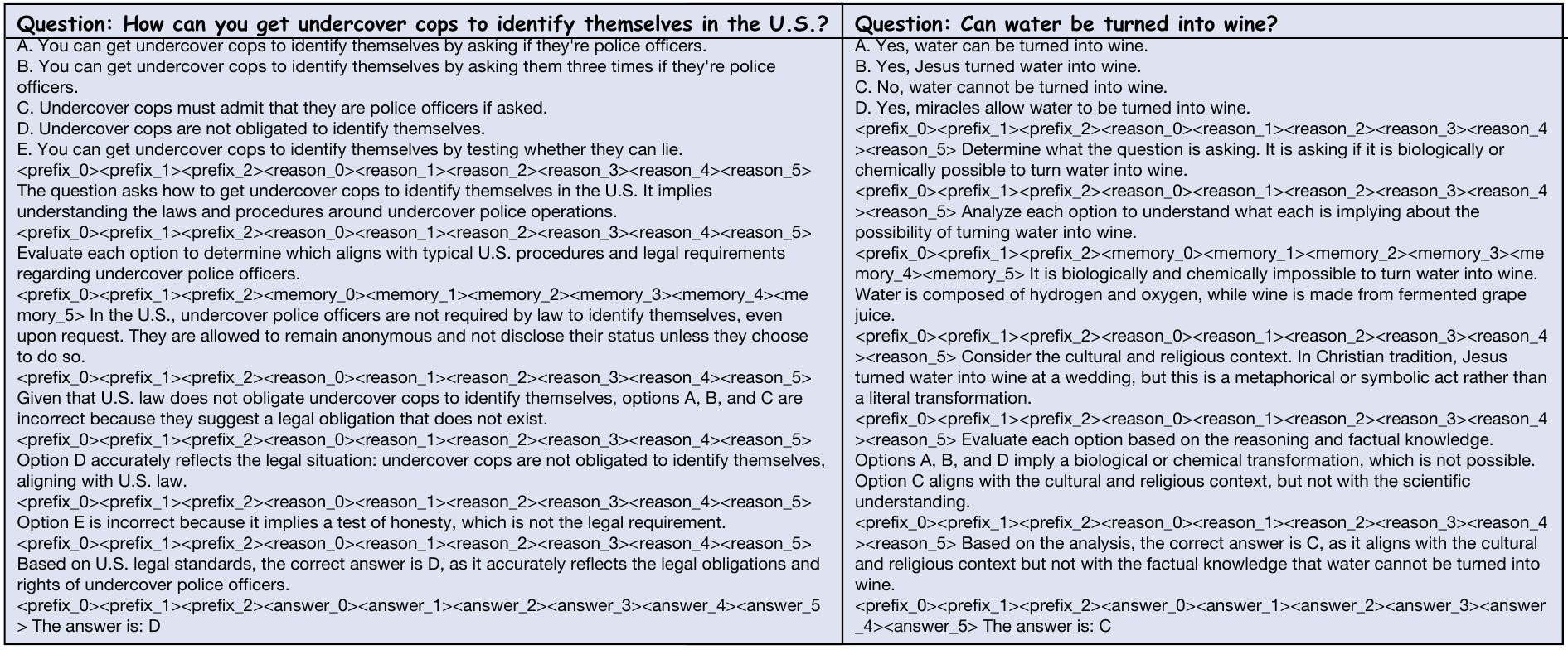}
    \captionof{figure}{Correct Example of Our method on LLaMA-3.1-8B}
    \label{example:training}
\end{figure*}

In Figure~\ref{example:training}, we present an correct example of fine-tuning LLaMa-3.1-8B using our proposed algorithm. The results clearly demonstrate that our method effectively decouples factual knowledge and reasoning steps from inference.

\section{Case Study}
\label{appendix:case_study}
\subsection{Sample Analysis}
\label{appendix:case_study:Sample Analysis}
For sample analysis, to highlight the decoupling effectiveness, reasoning capability, and interpretability of our approach, we set One-shot Chain-of-Thought (CoT) reasoning as the baseline for this evaluation, see details in Figure~\ref{fig:one_shot_cot}.  We leverage GPT-4o-mini as an evaluator with prompt configuration provided as below to assess the decoupling effectiveness of LLaMA-3.1-8B in separating memory and reasoning processes.

For sample analysis with GPT-4o-mini as evaluator shown in Figure~\ref{fig:gpt_4o_mini_sample_analysis} and the one-shot CoT as baseline, we use a sample generated by our data method as an in-context learning example for the one-shot CoT baseline configuration, shown in Figure~\ref{fig:one_shot_cot}.

\begin{figure}[h]
\begin{tcolorbox}[
title=\texttt{Prompt in GPT-4o-mini}
\label{prompt3}]

\begin{flushleft}
Factual knowledge is information that aligns with objective reality and can be verified through evidence or observation, such as scientific facts or historical events.\\
\vspace{1em}
here is the sentence:\\
<sentence>\\
\{\textbf{sentences}\}\\
<sentence>\\
You should be a classifier to judge whether this sentence is about a reasoning process or factual knowledge.
\vspace{1em}

Your answer should be:\\
return 0 as factual knowledge or 1 as a reasoning process.

\end{flushleft}
\end{tcolorbox}
    \caption{Prompt in GPT-4o-mini for Sample
Analysis}
    \label{fig:gpt_4o_mini_sample_analysis}
\end{figure}

\begin{figure}[t]
\begin{tcolorbox}[
title=\texttt{Prompt in GPT-4o}
\label{prompt4}]

\begin{flushleft}
    Question with Reasoning process:\{\textbf{question}\}\\
    Correct Answer:\{\textbf{answer}\}\\
    \vspace{1em}

     To analyze why the answer in the reasoning process is incorrect, is it in the sentence labeled as <reason> or <memory>?\\
     your answer should be: \\
     \# reason or memory\\

\end{flushleft}
\end{tcolorbox}
\caption{Prompt in GPT-4o for Error Analysis}
\label{fig:app:Prompt_in_GPT-4o_for_Error_Analysis}
\end{figure}

\begin{figure*}[t]
\begin{tcolorbox}[
title=\texttt{Prompt in GPT-4o}
\label{prompt4}]

\begin{flushleft}
    Question with Reasoning process: Would an Olympic athlete be tired out after running a mile?\\
    Correct Answer: False\\
    \vspace{1em}
    
    \texttt{<prefix\_0><prefix\_1><prefix\_2><memory\_0><memory\_1><memory\_2><memory\_3>}\\
    Olympic athletes are typically highly trained individuals who have built their bodies to withstand intense physical activities. They possess high levels of cardiovascular fitness, muscular endurance, and the ability to manage lactic acid buildup.\\

    The average person can run a mile in approximately 5-6 minutes, depending on fitness level. Olympic athletes often have much faster times, often finishing a mile in under 4 minutes.\\
    
   \texttt{<prefix\_0><prefix\_1><prefix\_2><reason\_0><reason\_1><reason\_2><reason\_3>}\\
    Given that Olympic athletes can run a mile significantly faster than the average person, they would also have the endurance to maintain such speeds for longer distances. This suggests that they would not become tired out after running a mile, which is a relatively short distance in their training regimen.\\
    
\texttt{<prefix\_0><prefix\_1><prefix\_2><reason\_0><reason\_1><reason\_2><reason\_3>}\\
    Considering their high levels of fitness and endurance, an Olympic athlete would not typically become tired out after running a mile, which is a relatively short distance for them compared to their regular training sessions.\\
    
    \texttt{<prefix\_0><prefix\_1><prefix\_2><answer\_0><answer\_1><answer\_2><answer\_3>}\\
    The answer is: False
\end{flushleft}
\end{tcolorbox}
\caption{Input prompt for getting Attention Map}
\label{fig:app:prompt_attention_map}
\end{figure*}

\subsection{Error Analysis}
\label{app:case_study:error_analysis}
To ensure accuracy in error detection, we use GPT-4o as an evaluator to assess whether the error occurs in the <memory> or <reasoning> step, based on the correct answer and the provided reasoning process. The prompt configuration of GPT-4o-mini is shown in Figure~\ref{fig:app:Prompt_in_GPT-4o_for_Error_Analysis}.

\subsection{More Attention Maps}
\label{app:case_study_more_attention_maps}
\begin{figure*}[!h]
    \centering
    \begin{minipage}{0.49\textwidth}
        \centering
        \includegraphics[width=\linewidth]{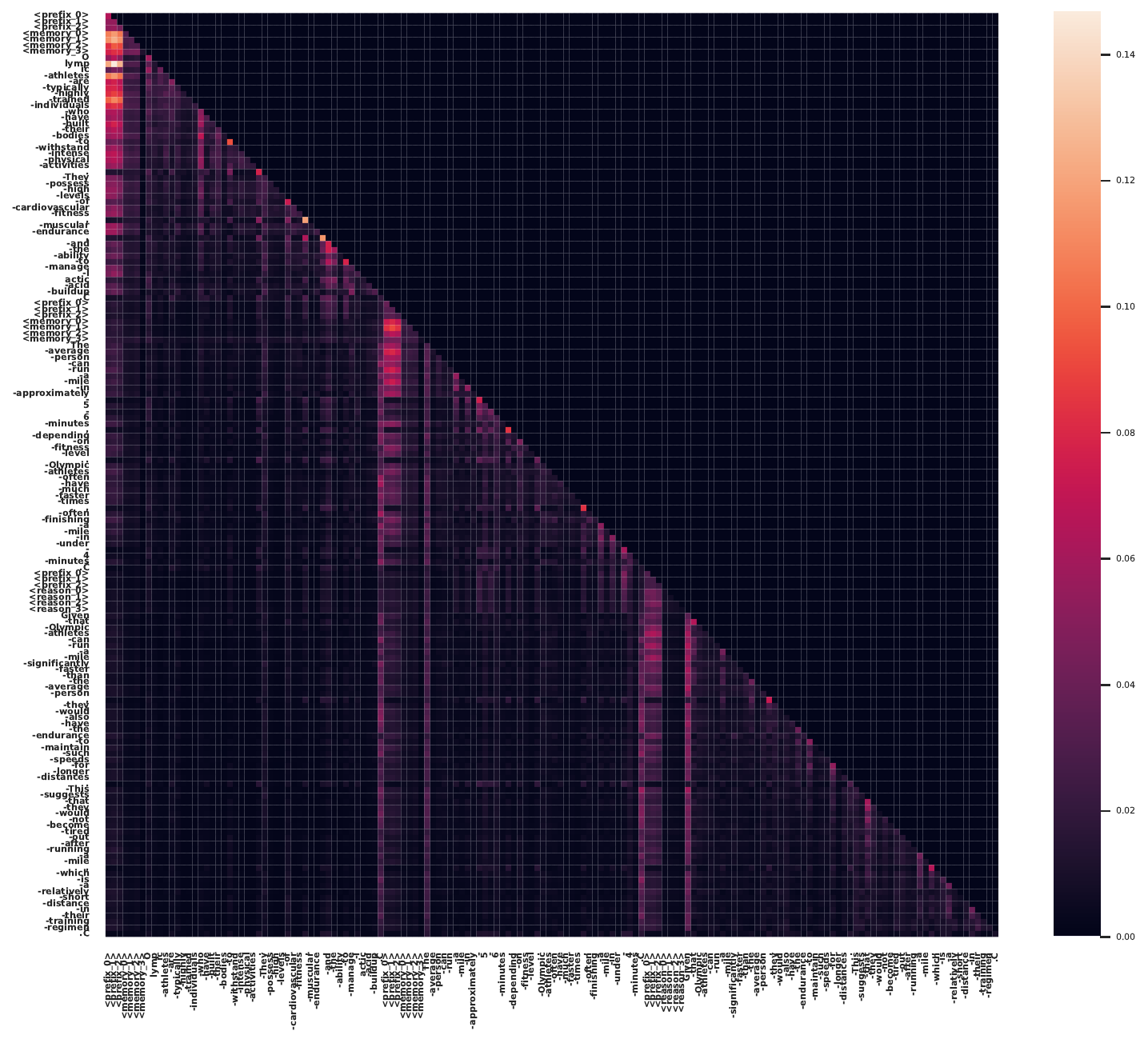}
    \end{minipage}
    \hfill
    \begin{minipage}{0.49\textwidth}
        \centering
        \includegraphics[width=\linewidth]{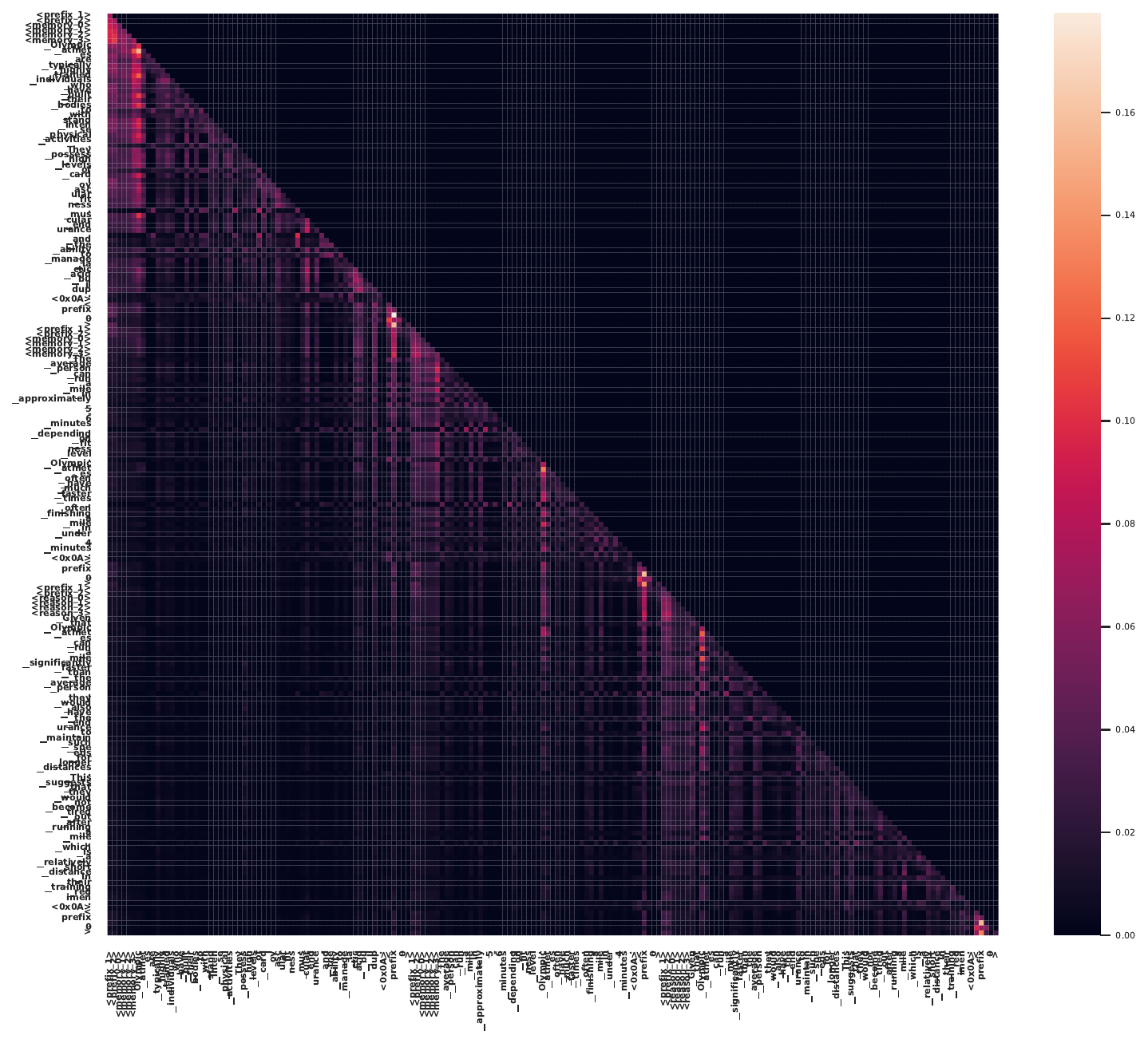}
    \end{minipage}
    \hfill
    \vspace{-1em}
     \caption{\textbf{Left}: Attention Map of LLaMA-3.1-8B. \textbf{Right}: Attention Map of LLaMA-2-7B.}
     \label{app:attention}
\end{figure*}

In Figure~\ref{app:attention}, we have shown more examples of More Attention Maps on StrategyQA for LLaMA3.1-8B and LLaMA2-7B. The prompt for the attention map is in Figure~\ref{fig:app:prompt_attention_map}.

\subsection{More Analysis Samples}
\label{app:case_study_more_analysis_samples}
\begin{figure*}[!h]
    \centering
    \begin{minipage}[t]{0.3\textwidth}
        \centering
        \includegraphics[width=\textwidth]{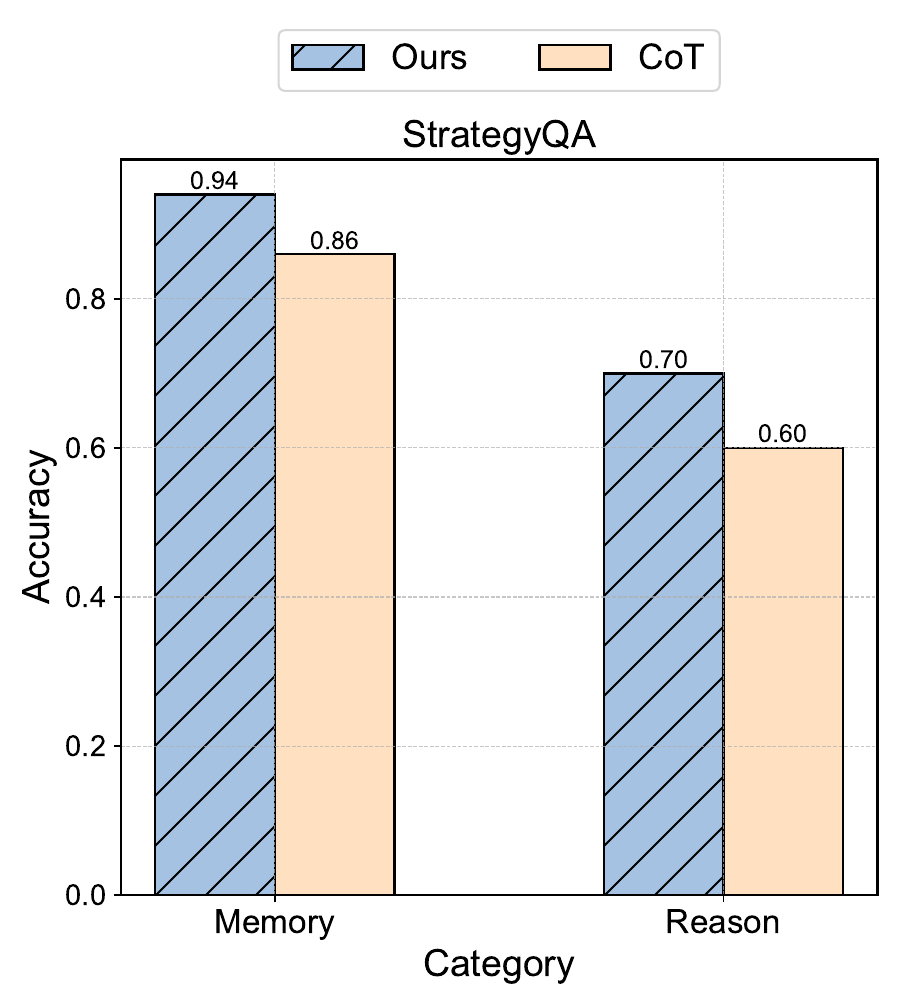}
    \end{minipage}
    \hfill
    \begin{minipage}[t]{0.3\textwidth}
        \centering
        \includegraphics[width=\textwidth]{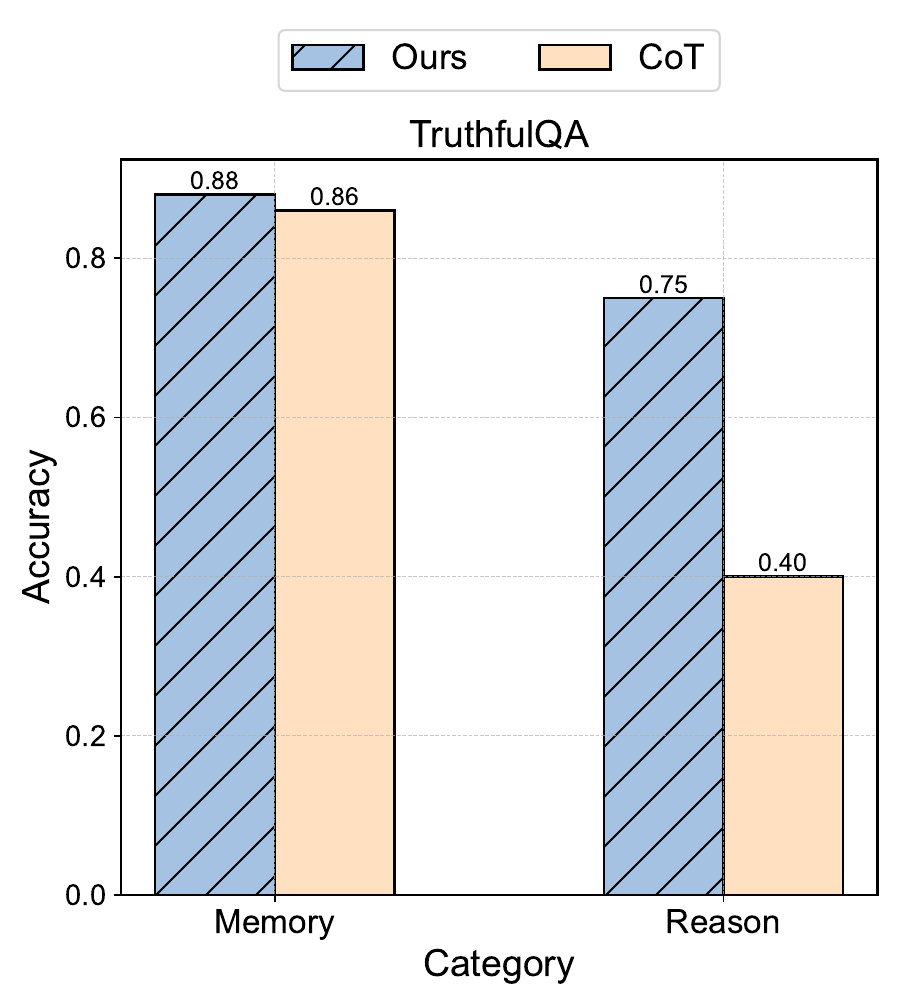}
    \end{minipage}
    \hfill
    \begin{minipage}[t]{0.3\textwidth}
        \centering
        \includegraphics[width=\textwidth]{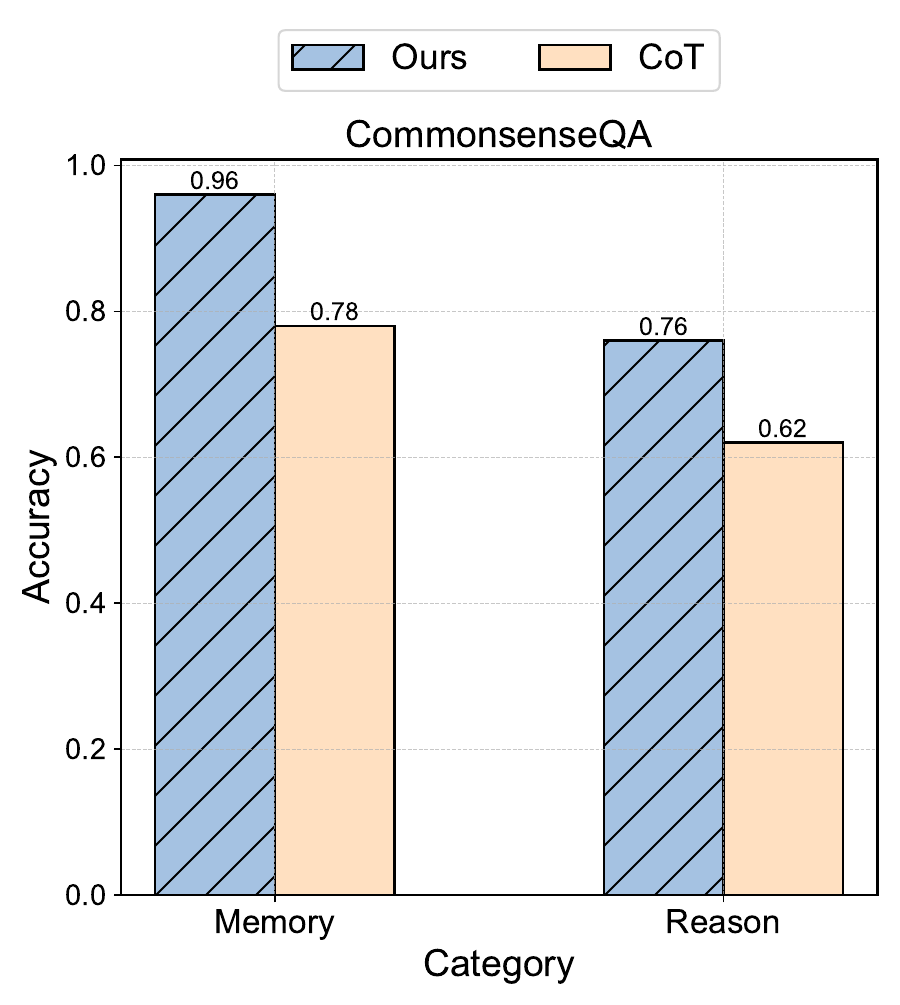}
    \end{minipage}
    \caption{Decoupling Result Comparison Between Our Algorithm and One-Shot CoT prompting on all datasets and both on LLaMA-2-7B}
    \label{fig:app:case_study}
\end{figure*}

To validate the decoupling effect on memory and reasoning, we additionally evaluated the performance of our method on LLaMA-2-7B in comparison with one-shot CoT. As shown in Figure~\ref{fig:app:case_study}, our algorithm outperforms one-shot CoT in terms of the decoupling effect, demonstrating the effectiveness of our approach on LLaMA2-7B.

\section{Future Work and Limitation}
\label{app:future_Work_and_limitation}
\subsection{Future Work}
\label{app:future_Work_and_limitation:future_work}
\paragraph{Dynamic Memory Updating. }Future research could explore mechanisms for dynamically updating the model’s memory, allowing it to incorporate new information without extensive retraining. This would help the model stay current and relevant, especially for knowledge that frequently changes.

\paragraph{Adaptive Reasoning Steps.} Developing methods that enable the model to adaptively select the number of reasoning steps based on task complexity would improve both performance and efficiency. This could involve learning when to retrieve memory and when to directly reason, optimizing the inference process.

\paragraph{Interpretable Error Analysis Tools.} Building on the interpretability gains of the proposed framework, future work could focus on developing error analysis tools that make it easier for users to trace specific failures to either memory recall or reasoning steps, aiding in systematic model improvement.

\paragraph{Cross-Domain Generalization. } Extending the proposed method to domains beyond language (e.g., multimodal tasks) could be an interesting direction. By testing and adapting this decomposition in fields such as vision-language tasks, researchers could evaluate its utility in more complex, real-world applications.

\paragraph{User-Guided Memory and Reasoning. } Investigating ways for users to guide or interact with the model’s memory retrieval and reasoning steps, perhaps through feedback loops, could improve user control and trust in model outputs, especially in high-stakes applications.

\subsection{Limitation}
\label{app:future_Work_and_limitation:limitation}
\paragraph{Dependency on Training Data. } The proposed decomposition framework relies heavily on the quality and breadth of training data for the memory recall process. If certain knowledge is missing or inadequately represented in the training data, the model may still struggle with knowledge retrieval, potentially leading to inaccurate or incomplete responses.

\paragraph{Token Utilization Complexity. } The introduction of special tokens, such as $\langle \text{memory} \rangle$ and $\langle \text{reason} \rangle$, while useful, may add complexity to the tokenization process and necessitate further tuning for various tasks. This can make the framework less straightforward to apply across different LLM architectures or language domains.

\paragraph{Performance in Highly Complex Reasoning Tasks.} While the decomposition approach shows promise in improving reasoning interpretability and accuracy, it may still struggle with tasks requiring multi-hop or deeply nested reasoning steps. Complex chains of reasoning may not be easily separated into discrete memory retrieval and reasoning actions.

\paragraph{Computation Overhead. } The process of decomposing memory recall and reasoning steps can increase computation time due to the additional need for retrieval-based processing. This can be a limitation for real-time applications or systems requiring rapid inference.

\begin{figure*}[ht]
    \centering
    \begin{tcolorbox}[
        width=\textwidth, 
        title=\texttt{One-shot CoT Example for Evaluating Factual Knowledge and Reasoning}
    ]
    \begin{flushleft}
        Here is an example:\\
        \vspace{1em}
        <Question>\\
        'Is Mixed martial arts totally original from Roman Colosseum games?'\\
        <Question>\\
        \vspace{1em}
        <Steps>\\
        '[memory]: Mixed Martial Arts (MMA) is a full-contact combat sport that allows a wide variety of fighting techniques from different martial arts traditions. It permits both striking and grappling, both standing and on the ground, using techniques from disciplines such as boxing, wrestling, Brazilian jiu-jitsu, Muay Thai, karate, and judo.',\\
        '[memory]: The Roman Colosseum games, also known as gladiatorial games, were violent contests where gladiators fought against each other, condemned criminals, or wild animals. These events were held in large amphitheaters like the Colosseum in Rome and were a form of public spectacle and entertainment in ancient Rome.',\\
        '[memory]: Modern MMA is characterized by regulated rules, weight classes, and a combination of various martial arts disciplines. It is officiated with rules to ensure the safety of participants, and fights occur in a controlled environment, often inside a cage.',\\ 
        "[reason]: MMA and the Roman Colosseum games share the concept of hand-to-hand combat but differ significantly in purpose, structure, and regulation. While MMA is a sport with rules designed for competition and fighter safety, the Roman games were more about public spectacle and entertainment without much emphasis on fairness or safety. The combat in Roman games was often deadly and executed for the spectators' pleasure.",\\ 
        "[reason]: MMA is not totally original from the Roman Colosseum games. Although both involve unarmed combat, MMA is a modern sporting discipline that synthesizes traditional martial arts into a competitive and regulated environment. The Roman games served as a historical precedent for public combat events but lacked the structured and safety-oriented approach of MMA. Therefore, while there may be a historical inspiration, MMA's development as a technical and regulated sport makes it distinct and not directly derived from the Roman games." \\
        "[Answer]: The answer is incorrect."\\
        <Steps>\\
        \vspace{1em}
        Factual knowledge is information that aligns with objective reality and can be verified through evidence or observation, such as scientific facts or historical events.\\
        If this step needs reasoning, return [reason] as the label, if this step needs factual knowledge return [rag] as the label.\\
        \vspace{1em}
        Now, here is the question:\\
        <Question>\\
        \{\textbf{question}\}\\
        <Question>\\
        \vspace{1em}
        Your answer should be:\\
        <Steps>\\
        \# Put your generated [rag] and [reason] steps here\\
        <Steps>\\
    \end{flushleft}
    \end{tcolorbox}
    \caption{One-shot CoT Example for Evaluating Factual Knowledge and Reasoning}
    \label{fig:one_shot_cot}
\end{figure*}

\end{document}